%% file: MoRM.tex
\documentclass{article}

\usepackage{microtype}
\usepackage{graphicx}
\usepackage{subfigure}
\usepackage{booktabs} % for professional tables

\usepackage{hyperref}

\usepackage[accepted]{icml2019}

\usepackage{amsmath}
\usepackage{amsfonts}
\usepackage{amssymb}
\usepackage{amsthm}
\usepackage{graphicx}
\usepackage{xr-hyper}
\usepackage{dsfont}
\usepackage{empheq}
\usepackage{enumitem}
\usepackage{bm}
\usepackage{bbm}
\usepackage{hhline}
\usepackage{mathtools}
\usepackage{stmaryrd}
\usepackage{cleveref}

\newtheorem{theorem}{{\bf Theorem}}

\newtheorem{proposition}{{\bf Proposition}}
\newtheorem{definition}{{\bf Definition}}

\newtheorem{lemma}{{\bf Lemma}}
\newtheorem{remark}{{\bf Remark}}

\def\mom{\hat{\theta}_\text{MoM}}
\def\morm{\bar{\theta}_\text{MoRM}}
\def\mormbis{\tilde{\theta}_\text{MoRM}}
\def\mou{\hat{\theta}_\text{MoU}(h)}
\def\moru{\bar{\theta}_\text{MoRU}(h)}
\def\moiu{\tilde{\theta}_\text{MoIU}(h)}
\def\sample{\mathcal{S}_n}

\newcommand\var[1]{\text{Var}\left({#1}\right)}

\icmltitlerunning{On Medians of Randomized (Pairwise) Means}

\begin{document}

\twocolumn[
\icmltitle{On Medians of (Randomized) Pairwise Means}



\icmlsetsymbol{equal}{*}

\begin{icmlauthorlist}
\icmlauthor{Pierre Laforgue}{tpt}
\icmlauthor{Stephan Cl\'emen\c{c}on}{tpt}
\icmlauthor{Patrice Bertail}{modalx}
\end{icmlauthorlist}

\icmlaffiliation{tpt}{LTCI, T\'el\'ecom Paris, Institut Polytechnique de Paris}
\icmlaffiliation{modalx}{Modal'X, UPL, Universit\'e Paris-Nanterre}

\icmlcorrespondingauthor{Pierre Laforgue}{\href{mailto:pierre.laforgue1@gmail.com}{pierre.laforgue1@gmail.com}}

\icmlkeywords{Machine Learning, ICML}

\vskip 0.3in
]



\printAffiliationsAndNotice{}  % leave blank if no need to mention equal contribution

\begin{abstract}
Tournament procedures, recently introduced in \citet{lugosi2016risk}, offer an appealing alternative, from a theoretical perspective at least, to the principle of \textit{Empirical Risk Minimization} in machine learning. Statistical learning by  Median-of-Means (MoM) basically consists in segmenting the training data into blocks of equal size and comparing the statistical performance of every pair of candidate decision rules on each data block: that with highest performance on the majority of the blocks is declared as the winner. In the context of nonparametric regression, functions having won all their duels have been shown to outperform empirical risk minimizers w.r.t. the mean squared error under minimal assumptions, while exhibiting robustness properties. It is the purpose of this paper to extend this approach, in order to address other learning problems in particular, for which the performance criterion takes the form of an expectation over pairs of observations rather than over one single observation, as may be the case in pairwise ranking, clustering or metric learning. Precisely, it is proved here that the bounds achieved by MoM are essentially conserved when the blocks are built by means of independent sampling without replacement schemes instead of a simple segmentation. These results are next extended to situations where the risk is related to a pairwise loss function and its empirical counterpart is of the form of a $U$-statistic.
%, it is also shown that
%proposed to build blocks of pairs of observations by sampling in the population of all training pairs, rather than forming all pairs based on data blocks preliminarily built (randomized or not), in order to reduce %significantly the variance of the resulting statistics.
Beyond theoretical results guaranteeing the performance of the learning/estimation methods proposed, some numerical experiments provide empirical evidence of their relevance in practice.
\end{abstract}

\input{1-Introduction}

\input{2-Preliminaries}

\input{3-Theory}

\input{4-Experiments}

\input{5-Conclusion}

\input{6-Proofs}

\clearpage
\bibliography{ref}
\bibliographystyle{icml2019}

\appendix
\input{Appendix}

\end{document}

%% file: 1-Introduction.tex
\section{Introduction}

In \citet{lugosi2016risk}, the concept of \textit{tournament procedure} for statistical learning has been introduced and analyzed in the context of nonparametric regression, one of the flagship problems of machine learning.
The task is to predict a real valued random variable (r.v.) $Y$ based on the observation of a random vector $X$ with marginal distribution $\mu(dx)$, taking its values in $\mathbb{R}^d$ with $d\geq 1$, say by means of a regression function $f:\mathbb{R}^d\rightarrow \mathbb{R}$ with minimum expected quadratic risk $\mathcal{R}(f)=\mathbb{E}[(Y-f(X))^2]$.
Statistical learning usually relies on a training dataset $\mathcal{S}_n=\{(X_1, Y_1),\; \ldots,\; (X_n, Y_n)\}$ formed of independent copies of the generic pair $(X,Y)$.
Following the \textit{Empirical Risk Minimization} (ERM) paradigm, one is encouraged to build predictive rules by minimizing an empirical version of the risk $\hat{\mathcal{R}}_n(f)=(1/n)\sum_{i=1}^n(y_i-f(x_i))^2$ over a class $\mathcal{F}$ of regression function candidates of controlled complexity (\textit{e.g.} of finite {\sc VC} dimension), while being rich enough to contain a reasonable approximant of the optimal regression function $f^*(x)=\mathbb{E}[Y\mid X=x]$: for any $f\in \mathcal{F}$, the risk excess $\mathcal{R}(f)-\mathcal{R}(f^*)$ is then equal to $\vert\vert f-f^*\vert\vert^2_{L_2(\mu)}=\mathbb{E}[(f(X)-f^*(X))^2]$.
A completely different learning strategy, recently proposed in \citet{lugosi2016risk}, consists in implementing a \textit{tournament procedure} based on the Median-of-Means (MoM) method (see \citet{nemirovsky1983problem}).

Precisely, the full dataset is first divided into $3$ subsamples of equal size.
For every pair of candidate functions $(f_1, f_2) \in \mathcal{F}^2$, the first step consists in computing the MoM estimator of the quantity $\|f_1 - f_2\|_{L_1(\mu)}:=\mathbb{E}[\vert f_1(X)-f_2(X) \vert]$ based on the first subsample: the latter being segmented into $K\geq 1$ subsets of equal size (approximately), $\|f_1 - f_2\|_{L_1(\mu)}$ is estimated by the median of the collection of estimators formed by its empirical versions computed from each of the $K$ sub-datasets.
When the MoM estimate is large enough, the match between $f_1$ and $f_2$ is allowed.
The rationale behind this approach is as follows: if one of the candidate, say $f_2$, is equal to $f^*$, and the quantity $\|f_1 - f^*\|_{L_1(\mu)}$ (which is less than $\|f_1 - f^*\|_{L_2(\mu)}=\sqrt{\mathcal{R}(f_1)- \mathcal{R}(f^*)}$) is large, so is its (robust) MoM estimate (much less sensitive to atypical values than sampling averages)  with high probability.
Therefore, $f^*$ is compared to distant candidates only, against which it should hopefully win its matches.
The second step consists in computing the MoM estimator of $\mathcal{R}(f_1) - \mathcal{R}(f_2)$ based on the second subsample for every \textit{distant enough} candidates $f_1$ and $f_2$.
If a candidate wins all its matches, it is kept for the third round.
As said before, $f^*$ should be part of this final pool, denoted by $H$.
Finally, matches involving all pairs of candidates in $H$ are computed, using a third MoM estimate on the third part of the data.
A champion winning again all its matches is either $f^*$ or has a small enough excess risk anyway.

It is the purpose of the present article to extend the MoM-based statistical learning methodology.
Firstly, we investigate the impact of randomization in the MoM technique: by randomization, it is meant that data subsets are built through sampling schemes, say simple random sampling without replacement (SRSWoR in abbreviated form) for simplicity, rather than partitioning.
Though introducing more variability in the procedure, we provide theoretical and empirical evidence that attractive properties of the original MoM method are essentially preserved by this more flexible variant (in particular, the number of blocks involved in this alternative procedure is arbitrary).
Secondly, we consider the application of the tournament approach to other statistical learning problems, namely those involving pairwise loss functions, like popular formulations of ranking, clustering or metric-learning.
In this setup, natural statistical versions of the risk of low variance take the form of $U$-statistics (of degree two), \textit{i.e.} averages over all pairs of observations, see \textit{e.g.} \citet{CLV08}.
In this situation, we propose to estimate the risk by the median of $U$-statistics computed from blocks obtained through data partitioning or sampling.
Results showing the accuracy of this strategy, referred to as \textit{Median of (Randomized) Pairwise Means} here, are established and application of this estimation technique to pairwise learning is next investigated from a theoretical perspective and generalization bounds are obtained.
The relevance of this approach is also supported by convincing illustrative numerical experiments.

The rest of the paper is organized as follows.
Section \ref{sec:preliminaries} briefly recalls the main ideas underlying the MoM procedure, its applications to robust machine learning as well as basic concepts pertaining to the theory of $U$-statistics/processes.
In section \ref{sec:main}, the variants of the MoM approach we propose are described at length and theoretical results establishing their statistical performance are stated.
Illustrative numerical experiments are displayed in section \ref{sec:num}, while proofs are deferred to the Appendix section. Some technical details and additional experimental results are postponed to the Supplementary Material (SM).

%% file: 2-Preliminaries.tex
\section{Background - Preliminaries}
\label{sec:preliminaries}

As a first go, we briefly describe the main ideas underlying the tournament procedure for robust machine learning, and next recall basic notions of the theory of $U$-statistics, as well as crucial results related to their efficient approximation.
Here and throughout, the indicator function of any event $\mathcal{E}$ is denoted by $\mathbb{I}\{\mathcal{E}\}$, the variance of any square integrable r.v. $Z$ by $\var{Z}$, the cardinality of any finite set $A$ by $\# A$.
If $(a_1,\; \ldots,\; a_n)\in\mathbb{R}^n$, the median (sometimes abbreviated med) of $a_1,\; \ldots,\; a_n$ is defined as $a_{\sigma((n+1)/2)}$ when $n$ is odd and $a_{\sigma(n/2)}$ otherwise, $\sigma$ denoting a permutation of $\{1,\; \ldots,\; n \}$ such that $a_{\sigma(1)}\leq \ldots\leq a_{\sigma(n)}$.
The floor and ceiling functions are denoted by $u\mapsto \lfloor u\rfloor$ and $u\mapsto \lceil u\rceil$.

\subsection{Medians of Means based Statistical Learning}\label{subsec:mom}

First introduced independently by \citet{nemirovsky1983problem}, \citet{jerrum1986random}, and \citet{alon1999space}, the Median-of-Means (MoM) is a mean estimator dedicated to real random variables.
It is now receiving a great deal of attention in the statistical learning literature, following in the footsteps of the results established in \citet{audibert2011robust}, \citet{catoni2012challenging}, where mean estimators are studied through the angle of their deviation probabilities, rather than on their traditional mean square errors, for robustness purpose.
Indeed, \citet{devroye2016sub} showed that the MoM provides an optimal $\delta$-dependent subgaussian mean estimator, under the sole assumption that a second order moment exists.
The MoM estimator has later been extended to random vectors, through different generalizations of the median \citep{minsker2015geometric, hsu2016loss, lugosi2017sub}.
In \citet{bubeck2013bandits}, it is used to design robust bandits strategies, while \citet{lerasle2011robust} and \citet{brownlees2015empirical} advocate minimizing a MoM, respectively Catoni, estimate of the risk, rather than performing ERM, to tackle different learning tasks.
More recently, \citet{lugosi2016risk} introduced a tournament strategy based on the MoM approach.

{\bf The MoM estimator.}
Let $\mathcal{S}_n = \{Z_1, \ldots, Z_n\}$ be a sample composed of $n\geq 1$ independent realizations of a square integrable real valued r.v. $Z$, with expectation $\theta$ and finite variance $\var{Z}=\sigma^2$.
Dividing $\mathcal{S}_n$ into $K$ disjoint blocks, each with same cardinality $B= \lfloor n / K \rfloor$, $\tilde{\theta}_k$ denotes the empirical mean based on the data lying in block $k$ for $k \le K$.
The MoM estimator $\mom$ of $\theta$ is then given by
$$
\mom = \text{median}(\tilde{\theta}_1,\; \ldots,\; \tilde{\theta}_K).
$$
It offers an appealing alternative to the sample mean $\hat{\theta}_n=(1/n)\sum_{i=1}^n Z_i$, much more robust, \textit{i.e.} less sensitive to the presence of atypical values in the sample.
Exponential concentration inequalities for the MoM estimator can be established in heavy tail situations, under the sole assumption that the $Z_i$'s are square integrable.
For any $\delta \in [e^{1 - n/2}, 1[$, choosing $K = \lceil \log(1 / \delta)\rceil$ and $B = \lfloor n / K \rfloor$, we have (see \textit{e.g.} \citet{devroye2016sub}, \citet{lugosi2016risk}):
\begin{equation}\label{eq:MoM_bound}
\mathbb{P}\left\{\left|\mom - \theta \right| > 2\sqrt{2}e\sigma\sqrt{\frac{1 + \log(1 / \delta)}{n}}\right\} \le \delta,
\end{equation}

{\bf The tournament procedure.}
Placing ourselves in the distribution-free regression framework, recalled in the Introduction section, it has been shown that, under appropriate complexity conditions and in a possibly non sub-Gaussian setup, the tournament procedure outputs a candidate $\hat{f}$ with optimal accuracy/confidence tradeoff, outperforming thus ERM in heavy-tail situations.
Namely, there exist $c,\;  c_0$, and $r > 0$ such that, with probability at least $1 - \exp(c_0 n \min\{1, r^2\})$, it holds both at the same time (see Theorem 2.10 in \citet{lugosi2016risk}):
\begin{equation*}
\|\hat{f} - f^*\|_{L_2} \le cr, \text{\quad and \quad} \mathcal{R}(\hat{f}) - \mathcal{R}(f^*) \le (cr)^2,
\end{equation*}

\subsection{Pairwise Means and $U$-Statistics}\label{subsec:Ustat}

Rather than the mean of an integrable r.v., suppose now that the quantity of interest is of the form $\theta(h)=\mathbb{E}[h(X_1,X_2)]$, where $X_1$ and $X_2$ are i.i.d. random vectors, taking their values in some measurable space $\mathcal{X}$ with distribution $F(dx)$ and $h:\mathcal{X} \times \mathcal{X} \rightarrow \mathbb{R}$ is a measurable mapping, square integrable w.r.t. $F\otimes F$.
For simplicity, we assume that $h(x_1,x_2)$ is symmetric (\textit{i.e.} $h(x_1,x_2)=h(x_2,x_1)$ for all $(x_1,x_2)\in \mathcal{X}^2$).
A natural estimator of the parameter $\theta(h)$ based on an i.i.d. sample $\mathcal{S}_n=\{X_1,\; \ldots,\; X_n\}$ drawn from $F$ is the average over all pairs
\begin{equation}\label{eq:Ustat}
U_n(h)=\frac{2}{n(n-1)}\sum_{1\leq i<j\leq n}h(X_i,X_j).
\end{equation}
The quantity \eqref{eq:Ustat} is known as the $U$-statistic of degree two\footnote{Let $d\geq n$ and $H:\mathcal{X}^d\rightarrow \mathbb{R}$ be measurable, square integrable with respect to $F^{\otimes k}$. The statistic $(n!/(n-d)!)\sum_{(i_1,\; \ldots,\; i_d)}H(X_{i_1},\; \ldots,\; X_{i_d})$, where the sum is taken over all $d$-tuples of $(1,\; \ldots,\; n)$, is a $U$-statistic of degree $d$.}, with kernel $h$, based on the sample $\mathcal{S}_n$.
One may refer to \citet{Lee90} for an account of the theory of $U$-statistics.
As may be shown by a Lehmann-Scheff\'e argument, it is the unbiased estimator of $\theta(h)$ with minimum variance.
Setting $h_1(X_1)=\mathbb{E}[h(X_1,X_2)\mid X_1]-\theta(h)$, $h_2(X_1,X_2)=h(X_1,X_2)-\theta(h)-h_1(X_1)-h_1(X_2)$, $\sigma_1^2(h)=\var{h_1(X_1)}$ and $\sigma^2_2(h)=\var{h_2(X_1,X_2)}$ and using the orthogonal decomposition (usually referred to as \textit{second Hoeffding decomposition}, see \citet{Hoeffding48})
\begin{align*}
U_n(h)-\theta(h) =&~\frac{2}{n}\sum_{i=1}^n h_1(X_i)\\
&+\frac{2}{n(n-1)}\sum_{1\le i < j \le n} h_2(X_i,X_j),
\end{align*}
one may easily see that
\begin{equation}\label{eq:var_Ustat}
\var{U_n(h)} = \frac{4\sigma_1^2(h)}{n}+\frac{2\sigma^2_2(h)}{n(n-1)}.
\end{equation}
Of course, an estimator of the parameter $\theta(h)$ taking the form of an i.i.d. average can be obtained by splitting the dataset into two halves and computing

$$
M_n(h)=\frac{1}{\lfloor n/2 \rfloor}\sum_{i=1}^{ \lfloor n/2 \rfloor}h(X_i,X_{i+\lfloor n/2 \rfloor}).
$$
One can check that its variance, $\text{Var}(M_n(h))=\sigma^2(h)/ \lfloor n/2 \rfloor$, with $\sigma^2(h)= \text{Var}(h(X_1, X_2) = 2\sigma^2_1(h)+\sigma^2_2(h)$, is however significantly larger than \eqref{eq:var_Ustat}.
Regarding the difficulty of the analysis of the fluctuations of \eqref{eq:Ustat} (uniformly over a class of kernels possibly), the reduced variance property has a price: the variables summed up being far from independent, linearization tricks (\textit{i.e.} Hajek/Hoeffding projection) are required to establish statistical guarantees for the minimization of $U$-statistics. Refer to \citet{CLV08} for further details.

\noindent {\bf Examples.} In machine learning, various empirical performance criteria are of the form of a $U$-statistic.

$\bullet$ In clustering, the goal is to find a partition $\mathcal{P}$ of the feature space $\mathcal{X}$ so that pairs of observations independently drawn from a certain distribution $F$ on $\mathcal{X}$ within a same cell of $\mathcal{P}$ are more similar w.r.t. a certain metric $D:\mathcal{X}^2\to \mathbb{R}_+$ than pairs lying in different cells. Based on an i.i.d. training sample $X_1,\; \ldots,\; X_n$, this leads to minimize the $U$-statistic, referred to as  \textit{empirical clustering risk}:
\begin{equation*}
\widehat{\mathcal{W}}_{n}(\mathcal{P})=\frac{2}{n(n-1)}\sum_{1\leq i<j \leq n}D(X_{i},X_{j})\cdot\Phi_{\mathcal{P}}(X_{i},X_{j}),
\end{equation*}
where $\Phi_{\mathcal{P}}(x,x')=\sum_{\mathcal{C}\in \mathcal{P}}\mathbb{I}\{(x,x')\in \mathcal{C}^2  \}$, over a class of partition candidates (see \citet{CLEM14}).

$\bullet$ In pairwise ranking, the objective is to learn from independent labeled data $(X_1,Y_1),\; \ldots,\; (X_n,Y_n)$ drawn as a generic random pair $(X,Y)$, where the real valued random label $Y$ is assigned to an object described by a r.v. $X$ taking its values in a measurable space $\mathcal{X}$, a ranking rule $r:\mathcal{X}^2\to \{-1,0,+1\}$ that permits to predict, among two objects $(X,Y)$ and $(X',Y')$ chosen at random, which one is preferred: $(X,Y)$ is preferred to $(X',Y')$ when $Y>Y'$ and, in this case, one would ideally have $r(X,X')=+1$, the rule $r$ being supposed anti-symmetric (\textit{i.e.} $r(x,x')=-r(x',x)$ for all $(x,x')\in \mathcal{X}^2$). This can be formulated as the problem of minimizing the $U$-statistic known as the \textit{empirical ranking risk} (see \citet{CLV05}) for a given loss function $\ell:\mathbb{R}\to \mathbb{R}_+$:
\begin{equation*}
\widehat{\mathcal{L}}_{n}(r)=\frac{2}{n(n-1)}\sum_{1\leq i<j \leq n}\ell \left(-r(X_{i},X_{j})\cdot(Y_{i}-Y_{j})\right).
\end{equation*}
Other examples of $U$-statistics are naturally involved in the formulation of metric/similarity-learning tasks, see \citet{BHS14} or \citet{VCB18}.
We also point out that the notion of $U$-statistic is much more general than that considered above: $U$-statistics of degree higher than two (\textit{i.e.} associated to kernels with more than two arguments) and based on more than one sample can be defined, see \textit{e.g.} Chapter 14 in \citet{van2000asymptotic} for further details.
The methods proposed and the results proved in this paper can be straightforwardly extended to this more general framework.

%% file: 3-Theory.tex
\section{Theoretical Results}\label{sec:main}

Mainly motivated by pairwise learning problems such as those mentioned in subsection \ref{subsec:Ustat}, it is the goal of this section to introduce and study several extensions of the MoM approach for robust statistical learning.

\input{3_1-Theory_morm}

\input{3_2-Theory_morustat}

\input{3_3-Discussion}

%% file: 3_1-Theory_morm.tex
\subsection{Medians of Randomized Means}\label{subsec:morm}

As a first go, we place ourselves in the setup of \Cref{subsec:mom}, and use the notation introduced therein.
But instead of dividing the dataset into disjoint blocks, an arbitrary number $K$ of blocks, of arbitrary size $B\leq n$, are now formed by sampling without replacement (SWoR), independently from $\sample$.
Each randomized data block $\mathcal{B}_k$, $k \le K$, is fully characterized by a random vector $\bm{\epsilon}_k = (\epsilon_{k, 1}, \ldots, \epsilon_{k, n})$, such that $\epsilon_{k, i}$ is equal to $1$ if the $i$-th observation has been selected in the $k$-th block, and to $0$ otherwise.
The $\bm{\epsilon}_{k}$'s are i.i.d. random vectors, uniformly distributed on the set $\Lambda_{n,B}=\{ \bm{\epsilon} \in\{0,1\}^n:\; \sum_{i=1}^n \epsilon_{i}=B  \}$ of cardinality $\binom{n}{B}$.
Equipped with this notation, the empirical mean computed from the $k$-th randomized block, for $k \le K$, can be written as $\bar{\theta}_k = (1/B)\sum_{i=1}^n \epsilon_{k,i}Z_i$.
The \textit{Median-of-Randomized Means} (MoRM) estimator $\morm$ is then given by
\begin{equation}\label{eq:MoRM}
\morm = \text{median}(\bar{\theta}_1, \ldots, \bar{\theta}_K).
\end{equation}
We point out that the number $K$ and size $B\leq n$ of the randomized blocks are arbitrary in the MoRM procedure, in contrast with the usual MoM approach, where $B=\lfloor n/K \rfloor$.
However, choices for $B$ and $K$ very similar to those leading to \eqref{eq:MoM_bound} lead to an analogous exponential bound, as revealed by \Cref{prop:morm}'s proof.
Because the randomized blocks are not independent, the argument used to establish \eqref{eq:MoM_bound} cannot be applied in a straightforward manner to investigate the accuracy of \eqref{eq:MoRM}.
Nevertheless, as can be seen by examining the proof of the result stated below, a concentration inequality can still be derived, using the conditional independence of the draws given $\mathcal{S}_n$, and a closed analytical form for the conditional probabilities $\mathbb{P}\{\vert \bar{\theta}_k-\theta\vert>\varepsilon \mid \mathcal{S}_n\}$, seen as $U$-statistics of degree $B$.
Refer to the Appendix for details.
\begin{proposition}\label{prop:morm}
Suppose that $Z_1,\; \ldots,\; Z_n$ are independent copies of a square integrable r.v. $Z$ with mean $\theta$ and variance $\sigma^2$.
Then, for any $\tau \in ]0, 1/2[$, for any $\delta \in [2e^{-8\tau^2n / 9}, 1[$, choosing $K =\lceil \log(2 / \delta)/(2(1/2 - \tau)^2)\rceil$ and $B = \lfloor 8 \tau^2 n/(9 \log(2/\delta)) \rfloor$, we have:

\begin{equation}\label{eq:morm_bound}
\mathbb{P}\left\{\left\vert \morm -\theta \right\vert > \frac{3 \sqrt{3}~\sigma}{2~\tau^{3/2}} \sqrt{\frac{\log(2/\delta)}{n}} \right\}\leq \delta.
\end{equation}
\end{proposition}

The bound stated above presents three main differences with \eqref{eq:MoM_bound}.
Recall first that the number $K$ of randomized blocks is completely arbitrary in the MoRM procedure and may even exceed $n$.
Consequently, it is always possible to build $\lceil \log(2 / \delta) / (2\left(1/2 - \tau\right)^2)\rceil$ blocks, and there is no restriction on the range of admissible confidence levels $\delta$ due to $K$.
Second, the size $B$ of the blocks can also be chosen completely arbitrarily in $\{1,\; \ldots,\; n \}$, and independently from $K$.
Proposition \ref{prop:morm} exhibits their respective dependence with respect to $\delta$ and $n$. Still, $B$ needs to be greater than $1$, which results in a restriction on the admissible $\delta$'s, such as specified.
Observe finally that $B$ never exceeds $n$.
Indeed for all $\tau \in ]0, 1/2[$, $8 \tau^2/(9 \log(2/\delta))$ does not exceeds $1$ as long as $\delta$ is lower than $2 \exp(-2/9) \approx 1.6$, which is always true.
Third, the proposed bound involves an additional parameter $\tau$, that can be arbitrarily chosen in $]0,1/2[$.
As may be revealed by examination of the proof, the choice of this extra parameter reflects a trade-off between the order of magnitude of $K$ and that of $B$: the larger $\tau$, the larger $K$, the larger the confidence range, the lower $B$ and the lower the constant in \eqref{eq:morm_bound} as well.
Since one can pick $K$ arbitrarily large, $\tau$ can be chosen as large as possible in $]0,1/2[$.
This way, one asymptotically achieves a $3\sqrt{6}$ constant factor, which is the same than that obtained in \citet{hsu2016loss} for a comparable confidence range.
However, the price of such an improvement is the construction of a higher number of blocks in practice (for a comparable number of blocks, the constant in \eqref{eq:morm_bound} becomes $27\sqrt{2}$).

\begin{remark}\label{rk:alt1}
{\sc (Alternative sampling schemes)}
We point out that other procedures than the SWoR scheme above (\textit{e.g.} Poisson/Bernoulli/Monte-Carlo sampling) can be considered to build blocks and estimates of the parameter $\theta$.
However, as discussed in the SM, the theoretical analysis of such variants is much more challenging, due to possible replications of the same original observation in a block.
\end{remark}

\begin{remark}\label{rk:vector_extension}
{\sc (Extension to random vectors)}
Among approaches extending MoMs to random vectors, that of \citet{minsker2015geometric} could be readily adapted to MoRM.
Indeed, once Lemma 2.1 therein has been applied, the sum of indicators can be bounded exactly as in \Cref{prop:morm}'s proof.
Computationally, MoRM only differs from MoM in the sampling, adding no difficulty, while multivariate medians can be computed efficiently \citep{hopkins2018sub}.
\end{remark}

\begin{remark}\label{rk:random_motivation}
{\sc (Randomization motivation)}
Theoretically, randomization being a natural alternative to data segmentation, it appeared interesting to study its impact on MoMs.
On the practical side, when performing a MoM Gradient Descent (GD), it is often needed shuffling the blocks at each step (see \textit{e.g}. Remark 5 in \citet{lecue2018robust}).
While this shuffling may seem artificial and ``ad hoc" in a MoM GD, it is already included and controlled with MoRM.
Finally, extending MoU to incomplete $U$-statistics like in subsection \ref{sec:extensions} first requires a MoM randomization's study.
\end{remark}

%% file: 3_2-Theory_morustat.tex
\subsection{Medians of (Randomized) $U$-statistics}
We now consider the situation described in subsection \ref{subsec:Ustat}, where the parameter of interest is $\theta(h)=\mathbb{E}[h(X_1,X_2)]$ and investigate the performance of two possible approaches for extending the MoM methodology.

\noindent {\bf Medians of $U$-statistics.}
The most straightforward way of extending the MoM approach is undoubtedly to form complete $U$-statistics based on $K$ subsamples corresponding to sets of indexes $I_1,\; \ldots,\; I_K$ of size $B=\lfloor n/K\rfloor$ built by segmenting the full sample, as originally proposed: for $k \in\{1,\; \ldots,\; K\}$,
\begin{equation*}
\hat{U}_{k}(h)=\frac{2}{B(B-1)}\sum_{(i,j)\in I_k^2,\; i<j}h(X_i,X_j).
\end{equation*}
The \textit{median of $U$-statistics} estimator (MoU in abbreviated form) of the parameter $\theta(h)$ is then defined as
\begin{equation*}
\mou = \text{median}(\hat{U}_{1}(h),\; \ldots,\; \hat{U}_{K}(h)).
\end{equation*}

The following result provides a bound analogous to \eqref{eq:MoM_bound}, revealing its accuracy.
\begin{proposition}\label{prop:U_stat_comp}
Let $\delta \in [e^{1 - 2n/9}, 1[$. Choosing $K = \lceil 9/2 \log(1 / \delta) \rceil$, we have with probability at least $1 - \delta$:
\begin{equation*}
\left\vert \mou -\theta(h)  \right\vert \leq \sqrt{  \frac{C_1\log\frac{1}{\delta}}{n} +  \frac{C_2\log^2(\frac{1}{\delta})}{n \left(2 n - 9\log\frac{1}{\delta}\right)}},
\end{equation*}
with $C_1 = 108\sigma_1^2(h)$ and $C_2 = 486\sigma_2^2(h)$.
\end{proposition}

We point out that another robust estimator $\mom(h)$ of $\theta$ could also have been obtained by applying the classic Mo(R)M methodology recalled in subsection \ref{subsec:mom} to the set of $\lfloor n/2 \rfloor$ i.i.d. observations $\{h(X_i,X_{i+\lfloor n/2 \rfloor}):\; 1\leq i\leq \lfloor n/2 \rfloor  \}$, see the discussion in subsection \ref{subsec:Ustat}.
In this context, we deduce from Eq. \eqref{eq:MoM_bound} with $K=\lceil \log(1/\delta) \rceil$ and $B=\lfloor \lfloor n/2\rfloor /K \rfloor$ that, for any $\delta\in [e^{1- \lfloor n/2\rfloor /2}, 1[$
\begin{equation*}
\left|\mom(h) - \theta(h) \right| \leq 2\sqrt{2}e\sigma(h)\sqrt{\frac{1 + \log(1 / \delta)}{\lfloor n/2\rfloor}}
\end{equation*}
with probability at least $1-\delta$.
The Mo(R)M strategies on independent pairs lead to constants respectively equal to $4e\sigma(h)$ and $6\sqrt{3}\sigma(h)$.
On the other hand, MoU reaches a $6\sqrt{3}\sigma_1(h)$ constant factor on its dominant term.
Recalling that $\sigma^2(h) = 2\sigma_1^2(h) + \sigma_2^2(h)$, beyond the $\sqrt{2}$ constant factor, MoU provides an improvement all the more significant that $\sigma_2^2(h)$ is large.
Another difference between the bounds is the restriction on $\delta$, which is looser in the MoU case.
This is due to the fact that the MoU estimator can possibly involve any pair of observations among the $n(n - 1)/2$ possible ones, in contrast to $\mom(h)$ that relies on the $\lfloor n / 2 \rfloor$ pairs set once and for all at the beginning only.
MoU however exhibits a more complex \textit{two rates} formula, but the second term being negligible the performance are not affected, as shall be confirmed empirically.

As suggested in subsection \ref{subsec:morm}, the data blocks used to compute the collection of $K$ $U$-statistics could be formed by means of a SRSWoR scheme.
Confidence bounds for such a median of randomized $U$-statistics estimator,  comparable to those achieved by the MoU estimator,  are stated below.

\noindent {\bf Medians of Randomized $U$-statistics.}
The alternative we propose consists in building an arbitrary number $K$ of data blocks $\mathcal{B}_1,\; \ldots,\; \mathcal{B}_K$ of size $B\leq n$ by means of a SRSWoR scheme, and, for each data block $\mathcal{B}_k$, forming all possible pairs of observations in order to compute
\begin{equation*}
\bar{U}_{k}(h)=\frac{1}{B(B-1)}\sum_{i<j}\bm{\epsilon}_{k,i}\bm{\epsilon}_{k,j}\cdot h(X_i,X_j),
\end{equation*}
where $\bm{\epsilon}_{k}$ denotes the random vector characterizing the $k-th$ randomized block, just like in subsection \ref{subsec:morm}.
Observe that, for all $k\in\{1,\; \ldots,\; K  \}$, we have: $\mathbb{E}[\bar{U}_{k}(h) \mid \mathcal{S}_n ]=U_n(h)$.
The \textit{Median of Randomized $U$-statistics} estimator of $\theta(h)$ is then defined as
\begin{equation}\label{eq:moru}
\moru = \text{median}(\bar{U}_{1}(h),\; \ldots,\; \bar{U}_{K}(h)).
\end{equation}
The following proposition establishes the accuracy of the estimator $\eqref{eq:moru}$, while emphasizing the advantages of the greater flexibility it offers when choosing $B$ and $K$.
\begin{proposition}\label{prop:U_stat_rand}
For any $\tau \in ]0, 1/2[$, for any $\delta \in [2e^{-8\tau^2n/9}, 1[$, choosing $K = \lceil \log(2/\delta) / (2(1/2 - \tau)^2) \rceil$ and $B = \lfloor 8 \tau^2 n / (9 \log(2/\delta)) \rfloor$, it holds w.p.a.l. $1 - \delta$:
\begin{equation*}
\left\vert \moru -\theta(h)  \right\vert \leq \sqrt{  \frac{C_1(\tau)\log\frac{2}{\delta}}{n} +  \frac{C_2(\tau)\log^2(\frac{2}{\delta})}{n \left(8 n - 9\log\frac{2}{\delta}\right)}},
\end{equation*}
with $C_1(\tau) = 27\sigma_1^2(h) /(2 \tau^3)$ and $C_2 = 243\sigma_2^2(h) / (4\tau^3)$.
\end{proposition}

\begin{remark}
Observe that \Cref{prop:U_stat_comp}'s constants (and bound) can be recovered asymptotically by letting $\tau \to 1/2$.
\end{remark}

\begin{remark}\label{rmk:general_ustat}
Propositions \ref{prop:U_stat_comp} and \ref{prop:U_stat_rand} remain valid for (multi-samples) $U$-statistics of arbitrary degree.
Refer to the SM for the general statements, and discussions about related approaches \citep{joly2016robust, minsker2018robust}.
\end{remark}

\subsection{MoU-based Pairwise Learning}

We now describe a version of the tournament method tailored to \textit{pairwise learning}.
Let $\mathcal{X}$ be a measurable space, $\mathcal{F} \subset \mathbb{R}^{\mathcal{X} \times \mathcal{X}}$ a class of decision rules, and $\ell:\mathcal{F}\times \mathcal{X}^2\to \mathbb{R}_+$ a given loss function.
The goal pursued here is to learn from $2n$ i.i.d. variables $X_1,\; \ldots,\; X_{2n}$ distributed as a generic r.v. $X$ valued in $\mathcal{X}$ a minimizer of the risk $\mathcal{R}(f)=\mathbb{E}[\ell(f,(X,X'))]$, where $X'$ denotes an independent copy of $X$.
In order to benefit from the standard tournament setting, we introduce the following notation: for every $f \in \mathcal{F}$, let $H_f(X, X') = \sqrt{\ell(f,(X,X'))}$ the kernel that maps every pair $(X, X')$ to its (square root) loss through $f$.
Let $\mathcal{H}_\mathcal{F} = \{H_f : f \in \mathcal{F}\}$.
It is easy to see that for all $f \in \mathcal{F}$, $\mathcal{R}(f) = \|H_f\|^2_{L_2(\mu)}$, and that if $f^*$ and $H^*_f$ denote respectively the $\mathcal{R}$ and $L_2$ minimizers over $\mathcal{F}$ and $\mathcal{H}_\mathcal{F}$, then $H^*_f = H_{f^*}$.
First, the dataset is split into $2$ subsamples $\mathcal{S}$ and $\mathcal{S}'$, each of size $n$.
Then, a distance oracle is used to allow matches to take place.
Namely, for any $f, g \in \mathcal{F}^2$, let $\Phi_\mathcal{S}(f, g)$ be a MoU estimate of $\|H_f - H_g\|_{L_1}$ built on $\mathcal{S}$. If $\mathcal{B}_1, \ldots, \mathcal{B}_K$ is a partition of $\mathcal{S}$, it reads:
\begin{equation*}
\Phi_\mathcal{S}(f, g) = \text{med}\left(\hat{U}_1|H_f - H_g|, \ldots, \hat{U}_K|H_f - H_g|\right).
\end{equation*}
If $\Phi_\mathcal{S}(f, g)$ is greater than $\beta r$, for $\beta$ and $r$ to be specified later, a match between $f$ and $g$ is allowed.
As shall be seen in the SM proofs, a MoRU estimate of $\|H_f - H_g\|_{L_1}$ could also have been used instead of a MoU.
The idea underlying the pairwise tournament is the same than that of the standard one: with high probability $\Phi_\mathcal{S}(f, g)$ is a good estimate of $\|H_f - H_g\|_{L_2}$, so that only distant candidates are allowed to confront.
And if $H_{f^*}$ is one of these two candidates, it should hopefully win its match against a distant challenger.
The nature of these matches is to be specified now.
For any $f, g \in \mathcal{F}^2$, let $\Psi_{\mathcal{S}'}(f, g)$ denote a MoU estimate of $\mathbb{E}[H_f^2 - H_g^2]$ built on $\mathcal{S}'$. With $\mathcal{B}'_1, \ldots, \mathcal{B}'_{K'}$ a partition of $\mathcal{S}'$, $\Psi_{\mathcal{S}'}(f, g)$ reads
\begin{equation*}
\Psi_{\mathcal{S}'}(f, g) = \text{med}\left(\hat{U}_1(H^2_f - H^2_g), \ldots, \hat{U}_{K'}(H^2_f - H^2_g)\right).
\end{equation*}
$f$ is declared winner of the match if $\Psi_{\mathcal{S}'}(f, g) \le 0$, \textit{i.e.} if $\sum_{i < j} \ell(f, (X_i, X_j))$ is lower than $\sum_{i < j} \ell(g, (X_i, X_j))$ on more than half of the blocks.
A candidate that has not lost a single match it has been allowed to participate in presents good generalization properties under mild assumptions, as revealed by the following theorem.

\begin{theorem}\label{thm:tournament}
Let $\mathcal{F}$ be a class of prediction functions, and $\ell$ a loss function such that $\mathcal{H}_\mathcal{F}$ is locally compact.
Assume that there exist $q > 2$ and $L > 1$ such that $\forall H_f \in \text{span}(\mathcal{H}_\mathcal{F}), \|H_f\|_{L_q} \le L\|H_f\|_{L_2}$.
Let $r^*$ (properly defined in the SM due to space limitation) that only depends on $f^*, L, q$, and the geometry of $\mathcal{H}_\mathcal{F}$ around $H_{f^*}$.
Set $r \ge 2r^*$.
Then there exist $c_0, c > 0$, and a procedure based on $X_1, \ldots, X_{2n}, L,$ and $r$ that selects a function $\hat{f} \in \mathcal{F}$ such that with probability at least $1 - \exp(c_0 n \min\{1, r^2\})$,
\begin{equation*}
\mathcal{R}(\hat{f}) - \mathcal{R}(f^*) \le cr.
\end{equation*}
\end{theorem}
\proof{
The proof is analogous to that of Theorem 2.11 in \citet{lugosi2016risk}, and sketched in the SM.
}
\begin{remark}
In pairwise learning, one seeks to minimize $\ell(f, X, X') = (\sqrt{\ell(f, X, X')} - 0)^2 = (H_f(X, X') - 0)^2$.
We almost recover the setting of \citet{lugosi2016risk}: quadratic loss, with $Y=0$, for the decision function $H_f$.
This is why any loss function $\ell$ can be considered, once technicalities induced by $U$-statistics are tackled.
The control obtained on $\|H_f - H^*_f\|_{L_2}$ then translates into a control on the excess risk of $f$ (see SM for further details).
\end{remark}

\begin{remark}
As discussed at length in \citet{lugosi2016risk}, computing the tournament winner is a nontrivial problem.
However, one could alternatively consider performing a tournament on an $\epsilon$-coverage of $\mathcal{F}$, while controlling the approximation error of this coverage.
\end{remark}

%% file: 3_3-Discussion.tex
\subsection{Discussion - Further Extensions}\label{sec:extensions}
The computation of the $U$-statistic \eqref{eq:Ustat} is expensive in the sense that it involves the summation of $\mathcal{O}(n^2)$ terms.
The concept of \textit{incomplete $U$-statistic}, see \citet{Blom76}, precisely permits to address this computational issue and achieve a trade-off between scalability and variance reduction.
In one of its simplest forms, it consists in selecting a subsample of size $M \geq  1$ by sampling with replacement in the set of all pairs of observations that can be formed from the original sample.
Setting
$
\Lambda=\{(i,j):\; 1\leq i<j\leq n \},
$
and denoting by $\{(i_1,j_1),\; \ldots,\; (i_M,j_M)\}\subset \Lambda$ the subsample drawn by Monte-Carlo, the incomplete version of the $U$-statistic \eqref{eq:Ustat} is: $\widetilde{U}_M(h)=(1/M)\sum_{m\leq M} h(X_{i_m},X_{j_m})$.
$\widetilde{U}_M(h)$ is directly an unbiased estimator of $\theta$ with variance
$$
\text{Var}\left(\widetilde{U}_M(h) \right)=\left(1-\frac{1}{M}\right)\text{Var}(U_n(h))+\frac{\sigma^2(h)}{M}.
$$
The difference between its variance and that of \eqref{eq:Ustat} vanishes as $M$ increases.
In contrast, when $M\leq \#\Lambda=n(n-1)/2$, the variance of a complete $U$-statistic based on a subsample of size $\lfloor \sqrt{M} \rfloor$, and thus on $\mathcal{O}(M)$ pairs just like $\widetilde{U}_M(h)$, is of order $\mathcal{O}(1/\sqrt{M})$.
Minimization of incomplete $U$-statistics has been investigated in \citet{CCB16} from the perspective of scalable statistical learning.
Hence, rather than sampling first observations and forming next pairs from data blocks in order to compute a collection of  \textit{complete $U$-statistics}, which the median is subsequently taken of, one could sample directly pairs of observations, compute alternatively estimates of drastically reduced variance and output a \textit{Median of Incomplete $U$-statistics}.
However, one faces significant difficulties when trying to analyze theoretically such a variant, as explained in the SM.

%% file: 4-Experiments.tex
\section{Numerical Experiments}\label{sec:num}

\begin{table*}[ht!]
\caption{Quadratic Risks for the Mean Estimation, $\delta = 0.001$}
\label{tab:MoRM_QR}
\vskip 0.15in
\begin{center}
\begin{small}
\begin{sc}
\begin{tabular}{lcccc}
\toprule
                                  & Normal $(0, 1)$       & Student $(3)$         & Log-normal $(0, 1)$   & Pareto $(3)$\\\midrule
MoM                               & 0.00149 $\pm$ 0.00218 & 0.00410 $\pm$ 0.00584 & 0.00697 $\pm$ 0.00948 & \textbf{1.02036 $\pm$ 0.06115}\\
$\text{MoRM}_\text{$1/6$, SWoR}$  & 0.01366 $\pm$ 0.01888 & 0.02947 $\pm$ 0.04452 & 0.06210 $\pm$ 0.07876 & 1.12256 $\pm$ 0.14970\\
$\text{MoRM}_\text{$3/10$, SWoR}$ & 0.00255 $\pm$ 0.00361 & 0.00602 $\pm$ 0.00868 & 0.01241 $\pm$ 0.01610 & 1.05458 $\pm$ 0.07041\\
$\text{MoRM}_\text{$9/20$, SWoR}$ & \textbf{0.00105 $\pm$ 0.00148} & \textbf{0.00264 $\pm$ 0.00372} & \textbf{0.00497 $\pm$ 0.00668} & 1.02802 $\pm$ 0.04903\\\bottomrule
\end{tabular}
\end{sc}
\end{small}
\end{center}
\vskip -0.1in
\end{table*}

\begin{table*}[ht!]
\caption{Quadratic Risks for the Variance Estimation, $\delta = 0.001$}
\label{tab:MoIU_QR}
\vskip 0.15in
\begin{center}
\begin{small}
\begin{sc}
\begin{tabular}{lcccc}
\toprule
                                  & Normal $(0, 1)$       & Student $(3)$         & Log-normal $(0, 1)$   & Pareto $(3)$\\\midrule
$\text{MoU}_\text{1/2; 1/2}$      & 0.00409 $\pm$ 0.00579 & 1.72618 $\pm$ 28.3563 & 2.61283 $\pm$ 23.5001 & 1.35748 $\pm$ 36.7998\\
$\text{MoU}_\text{Partition}$     & \textbf{0.00324 $\pm$ 0.00448} & \textbf{0.38242 $\pm$ 0.31934} & \textbf{1.62258 $\pm$ 1.41839} & \textbf{0.09300 $\pm$ 0.05650}\\
$\text{MoRU}_\text{SWoR}$         & 0.00504 $\pm$ 0.00705 & 0.51202 $\pm$ 3.88291 & 2.01399 $\pm$ 4.85311 & 0.09703 $\pm$ 0.07116\\\bottomrule
\end{tabular}
\end{sc}
\end{small}
\end{center}
\vskip -0.1in
\end{table*}

Here we display numerical results supporting the relevance of the MoM variants analyzed in the paper. Additional experiments are presented in the SM for completeness.

\noindent{\bf MoRM experiments.}
Considering inference of the expectation of four specified distributions (Gaussian, Student, Log-normal and Pareto), based on a sample of size $n = 1000$, seven estimators are compared below: standard MoM, and six MoRM estimators, related to different sampling schemes (SRSWoR, Monte-Carlo) or different values of the tuning parameter $\tau$.
Results are obtained through $5000$ replications of the estimation procedures.
Beyond the quadratic risk, accuracy of the estimators are assessed by means of deviation probabilities (see SM), \textit{i.e.} empirical quantiles for a geometrical grid of confidence levels $\delta$.
As highlighted above, $\tau = 1/6$ leads to (approximately) the same number of blocks as in the MoM procedure. However, MoRM usually select blocks of cardinality lower than $n / K$, so that the MoRM estimator with $\tau = 1/6$ uses less examples than MoM.
\Cref{prop:morm} exhibits a higher constant for MoRM in that case, and it is confirmed empirically here.
The choice $\tau = 3 / 10$ guarantees that the number of MoRM blocks multiplied by their cardinality is equal to $n$.
This way, MoRM uses as much samples as MoM. Nevertheless, the increased variability leads to a slightly lower performance in this case. Finally, $\tau = 9 / 20$ is chosen to be closer to $1/2$, as suggested by \eqref{eq:morm_bound}.
In this setting, the two constant factors are (almost) equal, and MoRM even empirically shows a systematic improvement compared to MoM.
Note that the quantile curves should be decreasing.
However, the estimators being $\delta$-dependent, different experiments are run for each value of $\delta$, and the rare little increases are due to this random effect.

\noindent{\bf Mo(R)U experiments.}
In these experiments assessing empirically the performance of the Mo(R)U methods, the parameter of interest is the variance (\textit{i.e.} $h(x, y) = (x - y)^2 / 2$) of the four laws used above.
Again, estimators are assessed through their quadratic risk and empirical quantiles.
A metric learning application is also proposed in the SM.

%% file: 5-Conclusion.tex
\section{Conclusion}\label{sec:concl}
In this paper, various extensions of the Medians-of-Means methodology, which tournament-based statistical learning techniques recently proposed in the literature to handle heavy-tailed situations rely on, have been investigated at length.
First, confidence bounds showing that accuracy can be fully preserved when data blocks are built through SRSWoR schemes rather than simple segmentation, giving more flexibility to the approach regarding the number of blocks and their size, are established.
Second, its application to estimation of pairwise expectations (\textit{i.e.} Medians-of-$U$-statistics) is studied in a valid theoretical framework, paving the way for the design of robust pairwise statistical learning techniques in clustering, ranking or similarity-learning tasks.

%% file: 6-Proofs.tex
\section*{Technical Proofs}

\subsection*{Proof of Proposition \ref{prop:morm}}

Let $\varepsilon>0$.
Just like in the classic argument used to prove \eqref{eq:MoM_bound}, observe that
$$
\left\{ \left|\morm - \theta\right| > \varepsilon \right\}\subset\left\{ \sum_{k=1}^K I_\varepsilon(\mathcal{B}_{\bm{\epsilon}_k})\geq K/2 \right\},
$$
where $I_\varepsilon(\mathcal{B}_{\bm{\epsilon}_k}) =\mathbb{I}\{\left\vert \bar{\theta}_k-\theta  \right\vert >\varepsilon \} \text{ for } k=1,\; \ldots,\; K$.
In order to benefit from the conditional independence of the blocks given the original sample $\sample$, we first condition upon $\sample$ and consider the variability induced by the $\bm{\epsilon}_k$'s only:
\begin{equation*}
\mathbb{P}\left\{\left|\morm - \theta\right| > \varepsilon \mid \sample \right\} \le \mathbb{P}\Bigg\{\sum_{k=1}^K \frac{I_\varepsilon(\mathcal{B}_{\bm{\epsilon}_k})}{K} \ge \frac{1}{2} \bigg| \sample \Bigg\}.
\end{equation*}
Now, the average $(1/K)\sum_{k=1}^K I_\varepsilon(\mathcal{B}_{\bm{\epsilon}_k})$ can be viewed as an approximation of the $U$-statistic of degree $B$ (refer to \citet{Lee90}), its conditional expectation given $\mathcal{S}_n$ being
\begin{equation*}
U_n^{\varepsilon}=\frac{1}{\binom{n}{B}}\sum_{\bm{\epsilon}\in \Lambda(n,B)}I_\varepsilon(\mathcal{B}_{\bm{\epsilon}}).
\end{equation*}
Denoting by $p^{\varepsilon}=\mathbb{E}[U_n^{\varepsilon}]=\mathbb{P}\{\vert  \bar{\theta}_1-\theta\vert >\varepsilon\}$ the expectation of the $I_\varepsilon(\mathcal{B}_{\bm{\epsilon}_k})$'s, we have $\forall \tau \in ]0, 1/2[$:
\begin{align}\label{eq:cut}
\mathbb{P}\big\{\left|\morm - \theta\right| > &\varepsilon \big\} \leq
\mathbb{P}_{\sample}\left\{ U_n^{\varepsilon}-p^{\varepsilon}  \ge \tau - p^{\varepsilon} \right\}\\
+~\mathbb{E}_{\sample}\Bigg[ \mathbb{P}_{\bm{\epsilon}}\Bigg\{ &\frac{1}{K}\sum_{k=1}^K I_\varepsilon(\mathcal{B}_{\bm{\epsilon}_k}) -U_n^{\varepsilon}  \ge \frac{1}{2} - \tau \bigg| \mathcal{S}_n\Bigg\}\Bigg].\nonumber
\end{align}
By virtue of Hoeffding inequality for i.i.d. averages (see \citet{Hoeffding63}) conditioned upon $\sample$, we have: $\forall t>0$,
\begin{equation}\label{eq:hoeff_condi}
\mathbb{P}_{\bm{\epsilon}}\left\{ \frac{1}{K}\sum_{k=1}^K I_\varepsilon(\mathcal{B}_{\bm{\epsilon}_k}) - U_n^{\varepsilon}  \geq t\bigg| \mathcal{S}_n  \right\}\leq \exp\left(-2Kt^2  \right).
\end{equation}
In addition, the version of Hoeffding inequality for $U$-statistics (\textit{cf} \citet{Hoeffding63}, see also Theorem A in Chapter 5 of \citet{Serfling}) yields: $\forall t>0$,
\begin{equation}\label{eq:hoeff_ustat}
\mathbb{P}_{\sample}\left\{ U_n^{\varepsilon}- p^{\varepsilon} \geq t \right\}\leq \exp\left(-2nt^2/B  \right).
\end{equation}
One may also show (see SM \ref{subsec:MoRM_var}) that $p^{\varepsilon} \leq  \frac{\sigma^2}{B\epsilon^2}$.
Combining this remark with equations \eqref{eq:cut}, \eqref{eq:hoeff_condi} and \eqref{eq:hoeff_ustat}, the deviation probability of $\morm$ can be bounded by
\begin{equation*}\label{eq:morm_split}
\exp\left(-2\frac{n}{B} \left(\tau - \frac{\sigma^2}{B\varepsilon^2}\right)^2\right) + \exp\left(-2K \left(\frac{1}{2} - \tau\right)^2\right).
\end{equation*}
Choosing $K = \lceil\log(2 / \delta)/(2(1/2 - \tau)^2\rceil$ and $B = \lfloor 8 \tau^2 n / (9 \log(2/\delta))\rfloor$ leads to the desired result.\qed

\subsection*{Proof of Proposition \ref{prop:U_stat_comp}}
The data blocks are built here by partitioning the original dataset into $K\leq n$ subsamples of size $B=\lfloor n/K \rfloor$.
Set $I_{\varepsilon, k}=\mathbb{I}\{ \vert \hat{U}_k(h)-\theta(h) \vert>\varepsilon \}$ for $k \in\{1,\; \ldots,\; K\}$.
Again, observe that $\mathbb{P}\big\{  \big| \mou(h) - \theta(h) \big| > \varepsilon \big\}$ is lower than
\begin{align*}
\mathbb{P}\Bigg\{&\frac{1}{K} \sum_{k=1}^K I_{\varepsilon, k} - q^\varepsilon \ge \frac{1}{2} - q^\varepsilon\Bigg\},
\end{align*}
where $q^\varepsilon=\mathbb{E}[ I_{\varepsilon, 1} ]=\mathbb{P}\{\vert \hat{U}_1(h)-\theta(h) \vert>\varepsilon  \}$.
By virtue of Chebyshev's inequality and equation \eqref{eq:var_Ustat}:
\begin{equation*}
q^\varepsilon \leq \frac{\text{Var}\big(\hat{U}_1(h)\big)}{\varepsilon^2} = \frac{1}{\varepsilon^2}\left(\frac{4\sigma_1^2(h)}{B}+\frac{2\sigma_2^2(h)}{B(B-1)}\right).
\end{equation*}
Using Hoeffding inequality, the deviation probability can thus be bounded by
\begin{equation*}
\exp\left(-2 K \left(\frac{1}{2} - \left(\frac{4\sigma_1^2(h)}{B\varepsilon^2}+\frac{2\sigma_2^2(h)}{B(B-1)\varepsilon^2}\right)\right)^2\right).
\end{equation*}

Choosing $K = \log(1/\delta)/(2(1/2 - \lambda)^2)$, $\lambda \in ]0, 1/2[$ gives:
\begin{equation*}
\varepsilon =  \sqrt{ C_1(\lambda) \frac{\log\frac{1}{\delta}}{n} + C_2(\lambda) \frac{\log^2(\frac{1}{\delta})}{n \left[2(\frac{1}{2} - \lambda)^2 n - \log\frac{1}{\delta}\right]}},
\end{equation*}
with $C_1(\lambda) = 2\sigma_1^2(h)/(\lambda (\frac{1}{2} - \lambda)^2)$ and $C_2(\lambda) = \sigma_2^2(h)/(\lambda(1/2 - \lambda)^2)$.
The optimal constant for the first and leading term is attained for $\lambda = 1/6$, which corresponds to $K = (9/2) \log(1/\delta)$ and gives:
\begin{equation*}
\varepsilon =  \sqrt{C_1 \frac{\log\frac{1}{\delta}}{n} + C_2 \frac{\log^2(\frac{1}{\delta})}{n \left(2 n - 9\log\frac{1}{\delta}\right)}},
\end{equation*}
with $C_1 = 108\sigma_1^2(h)$ and $C_2 = 486\sigma_2^2(h)$.
Finally, taking $\lceil K \rceil$ instead of $K$ does not change the result.\qed

\subsection*{Proof of Proposition \ref{prop:U_stat_rand}}
Here we consider the situation where the estimator is the median of $K$ randomized $U$-statistics, computed from data blocks built by means of independent SRSWoR schemes.
We set $\mathcal{I}_{\varepsilon}(\bm{\epsilon}_k) = \mathbb{I}\left\{ \vert \bar{U}_{k}(h)-\theta(h) \vert>\varepsilon  \right\}$.
For all $\tau \in ]0, 1/2[$, we have:
\begin{multline}\label{eq:dev_split}
\mathbb{P}\Big\{  \Big| \moru - \theta(h) \Big| > \varepsilon \Big\} \le \mathbb{P}\left\{ \frac{1}{K}\sum_{k=1}^K \mathcal{I}_{\varepsilon}(\bm{\epsilon}_k) \ge \frac{1}{2}\right\}\\
\leq  ~\mathbb{E}\left[\mathbb{P}\left\{\frac{1}{K}\sum_{k=1}^K \mathcal{I}_{\varepsilon}(\bm{\epsilon}_k) - W_{n}^{\varepsilon}\geq \frac{1}{2} - \tau  \mid \mathcal{S}_n\right\}  \right]\\
+\mathbb{P}\left\{ W_{n}^{\varepsilon}-\bar{q}^{\varepsilon} \geq \tau- \bar{q}^{\varepsilon} \right\},
\end{multline}
where we set
\begin{eqnarray*}
W_{n}^{\varepsilon} &= &\mathbb{E}\left[ \mathcal{I}_{\varepsilon}(\bm{\epsilon}_1) \mid \mathcal{S}_n\right] = \frac{1}{\binom{n}{B}}\sum_{\epsilon\in \Lambda_{n,B}} \mathcal{I}_{\varepsilon}(\epsilon),\\
\bar{q}^{\varepsilon} &=& \mathbb{E}[\mathcal{I}_{\varepsilon}(\bm{\epsilon}_1)]=\mathbb{P}\{\vert \bar{U}_1(h)-\theta(h) \vert>\varepsilon  \}.
\end{eqnarray*}
The conditional expectation $W_{n}^{\varepsilon} $, with mean $\bar{q}^{\varepsilon}$, is a $U$-statistic of degree $B$, so that Theorem A in Chapter 5 of \citet{Serfling} yields:
\begin{equation}\label{eq:sterfling}
\mathbb{P}\left\{ W_{n}^{\varepsilon}-\bar{q}^{\varepsilon} \geq \tau- \bar{q}^{\varepsilon} \right\}\leq \exp\left( -2\frac{n}{B}\left(\frac{1}{2}- \bar{q}^{\varepsilon} \right)^2 \right).
\end{equation}

One may also show (see SM \ref{subsec:MoRU_var}) that
\begin{equation}\label{eq:q_bound}
\bar{q}^{\varepsilon} \le \frac{1}{\varepsilon^2}\left(\frac{4\sigma_1^2(h)}{B} + \frac{2\sigma_2^2(h)}{B(B - 1)}\right).
\end{equation}

Combining \eqref{eq:dev_split}, standard Hoeffding's inequality conditioned on the data, as well as \eqref{eq:sterfling} together with \eqref{eq:q_bound} gives that the deviation of $\moru$ is upper bounded by
\begin{align}\label{eq:moru_split}
&\exp\left(-2K \left(\frac{1}{2} - \tau\right)^2\right)\\
+~&\exp\left( -2\frac{n}{B}\left(\tau- \frac{1}{\varepsilon^2}\left(\frac{4\sigma_1^2(h)}{B} + \frac{2\sigma_2^2(h)}{B(B - 1)}\right) \right)^2 \right).\nonumber
\end{align}

Choosing $K = \lceil \log(2/\delta) / (2(1/2 - \tau)^2) \rceil$ and $B = \lfloor 8 \tau^2 n / (9 \log(2/\delta)) \rfloor$ leads to the desired bound.\qed

%% file: Appendix.tex
\onecolumn
\section{Technical Details}\label{Appendix}

\subsection{Remark on Proposition \ref{prop:morm}'s Bound}

We point out that $K$ and $B$ are chosen so that \eqref{eq:hoeff_condi} and \eqref{eq:hoeff_ustat} are both bounded by $\delta/2$.
Given $\tau$, setting $B= \lfloor 2 (\tau - \lambda)^2 n/ \log(2/\delta)\rfloor$ for $\lambda \in ]0, \tau[$ yields a minimal constant for $\lambda = \tau/3$.
Interestingly, $B$ involves a \textit{floor} function, even if it is not constrained by $K$.
An interpretation is that building large blocks increases the risk of selecting extreme values, and thus of deteriorating the performance.

\subsection{Variance Computations}

\subsubsection{MoRM}\label{subsec:MoRM_var}

By virtue of Chebyshev's inequality, one gets:
\begin{equation*}
p^\varepsilon \le \frac{\mathbb{E}\left[(\bar{\theta}_1 - \theta)^2 \right]}{\varepsilon^2} = \frac{\mathbb{E}_{\sample}\left[\mathbb{E}\left[(\bar{\theta}_1 - \theta)^2 \mid \sample\right]\right]}{\varepsilon^2}.
\end{equation*}
Observing that $\mathbb{E}[\bar{\theta}_1 | \sample] = \hat{\theta}_n$ and that
\begin{equation*}
\mathbb{E}\left[(\bar{\theta}_1 - \theta)^2 | \sample\right] = \var{\bar{\theta}_1 \mid \sample}+(\hat{\theta}_n-\theta)^2 = (\hat{\theta}_n-\theta)^2+\frac{1}{B}\frac{n-B}{n} \hat{\sigma}^2_n,
\end{equation*}
where $\hat{\sigma}^2_n=(1/(n-1))\sum_{i=1}^n(Z_i-\hat{\theta}_n)^2$, we deduce that
\begin{equation*}
p^{\varepsilon} \le \left(\frac{1}{n}+\frac{n-B}{nB} \right)\frac{\sigma^2}{\varepsilon^2} = \frac{\sigma^2}{B\epsilon^2}.
\end{equation*}
\qed

\subsubsection{MoRU}\label{subsec:MoRU_var}

Observe first that
\begin{equation}\label{eq:decomp_var}
\text{Var}\left( \bar{U}_1(h) \right)= \mathbb{E}\left[\text{Var}( \bar{U}_1(h) \mid \mathcal{S}_n)  \right] + \text{Var}\left( \mathbb{E}\left[\bar{U}_1(h) \mid \mathcal{S}_n\right] \right).
\end{equation}

Recall that $\mathbb{E}[\bar{U}_1(h) \mid \mathcal{S}_n]=U_n(h)$, so that
\begin{equation}\label{eq:var_exp_cond}
\text{Var}\left( \mathbb{E}\left[\bar{U}_1(h) \mid \mathcal{S}_n\right] \right)=  \frac{4\sigma_1^2(h)}{n}+\frac{2\sigma^2_2(h)}{n(n-1)}.
\end{equation}
In addition, we have, for $B\geq 4$,
\begin{equation*}
\text{Var}( \bar{U}_1(h) \mid \mathcal{S}_n) = \frac{4}{B^2(B-1)^2}\sum_{i<j}h^2(X_i,X_j)\text{Var}(\epsilon_{1,i}\epsilon_{1,j})~~~+ \hspace{-0.3cm}\sum_{\substack{i<j,\; k<l\\(i,j)\neq (k,l)}} \hspace{-0.4cm}\text{Cov}(\epsilon_{1,i}\epsilon_{1,j},\; \epsilon_{1,l}\epsilon_{1,k})h(X_i,X_j)h(X_k,X_l).
\end{equation*}
Let $i\neq j$, one may check that
$$
\text{Var}(\epsilon_{1,i}\epsilon_{1,j})=\frac{B(B-1)(n - B)(n + B - 1)}{n^2(n-1)^2}.
$$
And, for any $k\neq l$, we have
\begin{equation*}
\text{Cov}(\epsilon_{1,i}\epsilon_{1,j},\; \epsilon_{1,l}\epsilon_{1,k})= -\frac{B(B-1)}{n(n-1)} \frac{(n-B)(4nB-6n-6B+6)}{n(n-1)(n-2)(n-3)}
\end{equation*}
when $\{i,j \}\cap\{k,l  \}=\emptyset$, as well as
\begin{equation*}
\text{Cov}(\epsilon_{1,i}\epsilon_{1,j},\; \epsilon_{1,i}\epsilon_{1,k})= \frac{B(B-1)}{n(n-1)}\frac{(n-B)(nB-2n-2B+2)}{n(n-1)(n-2)}
\end{equation*}

when $k\neq j$ and $k\neq i$. Hence, observing that $\mathbb{E}[h(X_1,X_2)h(X_1,X_3)]=\sigma^2_1(h)+\theta^2(h)$, we obtain:
\begin{align}\label{eq:exp_var_cond}
\mathbb{E}\left[\text{Var}( \bar{U}_1(h) \mid \mathcal{S}_n)  \right] &=~\frac{2(n-B)(n + B - 1)}{n(n - 1)B(B-1)}\left( \sigma^2(h)+\theta^2(h) \right) ~~ - ~~ \frac{(n-B)(4nB-6n-6B+6)}{n(n-1)B(B-1)}\theta^2(h)\nonumber\\
&\quad+~~\frac{4(n-B)(nB-2n-2B+2)}{n(n-1)B(B-1)}(\sigma^2_1(h)+\theta^2(h)).
\end{align}
Combining \eqref{eq:decomp_var}, \eqref{eq:var_exp_cond} and \eqref{eq:exp_var_cond}, we get:
\begin{align*}
\text{V}\text{ar}\left( \bar{U}_1(h) \right) &=~\frac{4\sigma_1^2(h)}{n} ~~ + ~~ \frac{2\sigma^2_2(h)}{n(n-1)} ~~ + ~~ \frac{2(n-B)(n + B - 1)}{n(n - 1)B(B-1)}\left( 2\sigma_1^2(h) + \sigma_2^2 + \theta^2(h) \right)\\
&\quad - ~~ \frac{(n-B)(4nB-6n-6B+6)}{n(n-1)B(B-1)}\theta^2(h) ~~ + ~~ \frac{4(n-B)(nB-2n-2B+2)}{n(n-1)B(B-1)}(\sigma^2_1(h)+\theta^2(h)),\\
\text{V}\text{ar}\left( \bar{U}_1(h) \right) &=~\frac{4\sigma_1^2(h)}{B} + \frac{2\sigma_2^2(h)}{B(B - 1)}.
\end{align*}

Chebyshev inequality permits to conclude.\qed

\subsection{Remark on the Term $\log(2/\delta)$ in the Rate Bounds}
In all results related to randomized versions (namely \Cref{prop:morm} and \Cref{prop:U_stat_rand}), the term $\log(2/\delta)$ appears, instead of $\log(1/\delta)$.
We point out that this limitation can be easily overcome by means of a more careful analysis in \eqref{eq:morm_split} and \eqref{eq:moru_split}.
Indeed, $K$ and $B$ have been chosen so that both exponential terms are equal to $\delta / 2$, but one could of course consider splitting the two terms into $(1 - \kappa) \delta$ and $\kappa \delta$ for any $\kappa \in ]0, 1[$. This, way, choosing
$$
K = \left\lceil \log\left(\frac{1}{(1 - \kappa)\delta}\right) / (2(1/2 - \tau)^2)\right\rceil \text{ and } B = \left\lfloor 8\tau^2n / (9 \log\left(\frac{1}{\kappa \delta}\right))\right\rfloor
$$
leads to $\log(1/\kappa \delta)$ instead.

\subsection{Extension to Generalized $U$-statistics}

As noticed in Remark \ref{rmk:general_ustat}, Propositions \ref{prop:U_stat_comp} and \ref{prop:U_stat_rand} have been established for $U$-statistics of degree $2$, but remain valid for generalized ones.
For clarity, we recall the definition of generalized $U$-statistics.
An excellent account of properties and asymptotic theory of U-statistics can be found in \citet{Lee90}.
\begin{definition}\label{def:gen_ustat}
Let $T \ge 1$ and $(d_1, \ldots, d_T) \in {\mathbb{N}^*}^T$.
Let $\bm{X}_{\{1, \ldots, n_t\}} = (X_1^{(t)}, \ldots, X_{n_t}^{(t)}),~1 \le t \le T$, be $T$ independent samples of sizes $n_t \ge d_t$ and composed of i.i.d. random variables taking their values in some measurable spaces $\mathcal{X}_t$ with distribution $F_t(dx)$ respectively.
Let $H: \mathcal{X}_1^{d_1} \times \ldots \times \mathcal{X}_T^{d_T} \rightarrow \mathbb{R}$ be a measurable function, square integrable with respect to the probability distribution $\mu = F_1 ^{\otimes d_1} \otimes \ldots \otimes F_T^{\otimes d_T}$.
Assume in addition (without loss of generality) that $H(\bm{x}^{(1)}, \ldots, \bm{x}^{(T)})$ is symmetric within each block of argument $\bm{x}^{(t)}$ valued in $\mathcal{X}_t^{d_t},~1 \le t \le T$.
The generalized (or $T$-sample) $U$-statistic of degrees $(d_1, \ldots, d_T)$ with kernel $H$ is then defined as
\begin{equation*}
U_{\bm{n}}(H) = \frac{1}{\prod_{t=1}^T \binom{n_t}{d_t}} \sum_{I_1} \ldots \sum_{I_T} H(\bm{X}_{I_1}^{(1)}, \ldots, \bm{X}_{I_T}^{(T)}),
\end{equation*}
where the symbol $\sum_{I_t}$ refers to the summation over all $\binom{n_t}{d_t}$ subsets $\bm{X}_{I_t}^{(t)} = (X_{i_1}^{(t)}, \ldots, X_{i_{d_t}}^{(t)})$ related to a set $I_t$ of $d_t$ indexes $1 \le i_1 < \ldots < i_{d_t} \le n_t$ and $\bm{n} = (n_1, \ldots, n_T)$.
\end{definition}

Within this framework, we aim at estimating $\theta(h) = \mathbb{E}\left[H(X_1^{(1)}, \ldots, X_{d_1}^{(1)}, \ldots, X_1^{(T)}, \ldots, X_{d_T}^{(T)})\right]$, and an analog of the standard MoM estimator for generalized $U$-statistics can be defined as follows.
\begin{definition}\label{def:mogu}
With the notation introduced in Definition \ref{def:gen_ustat}, let $1\leq K \le \min_t n_t / (d_t + 1)$.
Partition each sample $\bm{X}^{(t)}$ into $K$ blocks $B_1^{(t)}, \ldots, B_K^{(t)}$ of sizes $\lfloor n_t / K \rfloor$.
Compute $\hat{\theta}_k$ the complete $U$-statistics based on $B_k^{(1)}, \ldots, B_k^{(T)}$ for $1 \le k \le K$.
The Median-of-Generalized-$U$-statistics if then given by $\hat{\theta}_\text{MoGU} = \text{median}(\hat{\theta}_1, \ldots, \hat{\theta}_K)$.
\end{definition}

Please note that this estimator is very different from that considered in \citet{minsker2018robust}.
Robust versions of $U$-statistics have already been considered in the literature (for the purpose of covariance estimation, rather than the design of statistical learning methods), from a completely different angle however.
The $U$-statistic is viewed as a $M$-estimator minimizing a criterion involving the quadratic loss, and the proposed estimator is the $M$-estimator solving the same criterion except that a different loss function is used.
This loss function is designed to induce robustness, while being close enough to the square loss to derive guarantees.

The definition above is closer to that given in \citet{joly2016robust}.
Indeed, in the particular setting of a $1$-sample $U$-statistic of degree $d_1$, Definition \ref{def:mogu} coincides with the \emph{diagonal blocks} estimate mentioned on page 5 of \citet{joly2016robust}.
However, as noticed therein, this estimator only considers a small fraction of possible $d_t$-tuples, namely those whose items are all in the block $B_k^{(t)}$.
In order to overcome this limitations, an alternative strategy involving decoupled $U$-statistics is adopted in \citet{joly2016robust}.
The approach pursued here is rather to consider randomized $U$-statistics, which is another way to introduce variability into the tuples considered to build the estimator.

\begin{proposition}
Using the notation of Definitions \ref{def:gen_ustat} and \ref{def:mogu}, let $n_\text{min} = \min_t n_t$, and $\underline{n}, \underline{d}$ such that $\underline{n}/\underline{d} = \min_t n_t/d_t$.
Let $\delta \in [e^{1 - 2\underline{n}/9\underline{d}}, 1[$. Choosing $K = \lceil 9/2 \log(1 / \delta) \rceil$, we have with probability at least $1 - \delta$:
\begin{equation*}
\left\vert \hat{\theta}_\text{MoGU} -\theta(h)  \right\vert \leq \sqrt{  \frac{C_1\log\frac{1}{\delta}}{n_\text{min}} +  \frac{C_2\log^2(\frac{1}{\delta})}{n_\text{min} \left(2 n_\text{min} - 9\log\frac{1}{\delta}\right)}},
\end{equation*}
with $C_1 = 108\sigma_1^2(h)$ and $C_2 = 486\sigma_2^2(h)$.
\end{proposition}

\begin{proposition}
Now let $\hat{\theta}_\text{MoRGU}$ (Median-of-Randomized-Generalized-$U$-statistics) be the estimator of $\theta(h)$ such that the blocks $B_k^{(t)}$ are no longer partitions of the samples $\bm{X}^{(t)}$, but rather drawn from SWoR.
For any $\tau \in ]0, 1/2[$, for any $\delta \in [2e^{-8\tau^2\underline{n}/9\underline{d}}, 1[$, choosing $K = \lceil \log(2/\delta) / (2(1/2 - \tau)^2) \rceil$ and $B_t = \lfloor 8 \tau^2 n_t / (9 \log(2/\delta)) \rfloor$, it holds with probability larger than $1 - \delta$:
\begin{equation*}
\left\vert \hat{\theta}_\text{MoRGU} -\theta(h)  \right\vert \leq \sqrt{  \frac{C_1(\tau)\log\frac{2}{\delta}}{n_\text{min}} +  \frac{C_2(\tau)\log^2(\frac{2}{\delta})}{n_\text{min} \left(8 n_\text{min} - 9\log\frac{2}{\delta}\right)}},
\end{equation*}
with $C_1(\tau) = 27\sigma_1^2(h) /(2 \tau^3)$ and $C_2 = 243\sigma_2^2(h) / (4\tau^3)$.
\end{proposition}

\par{\bf Proofs.}{
The proofs are analogous to that of Propositions \ref{prop:U_stat_comp} and \ref{prop:U_stat_rand}, except that concentration results for generalized $U$-statistics are used (see \textit{e.g.} \citet{Hoeffding63}).
}

\subsection{Proof of \Cref{thm:tournament} (sketch of)}

The proof follows the path of Theorem 2.11's proof in \citet{lugosi2016risk}, with adjustments every time $U$-statistics are involved instead of standard means. The first one deals with the constants involved in the propositions.

\begin{definition}
Let $\lambda_\mathbb{Q}(\kappa, \eta, h)$ and $\lambda_\mathbb{M}(\kappa, \eta, h)$ defined as in \citet{lugosi2016risk} (see Definitions 2.2 and 2.3 therein).
\end{definition}

\begin{definition}\label{def:changes}
A difference however occurs on $r_E(\kappa, h)$ and $\bar{r}_\mathbb{M}(\kappa, h)$. Indeed, let
\begin{equation*}
r_E(\kappa, h) = \inf\left\{r: \mathbb{E} \sup_{u \in \mathcal{F}_{h, r}} \left| \sqrt{\frac{2}{B(B - 1)}} \sum_{i < j} \sigma_{i, j}~u(X_i, X_j)  \le \kappa \sqrt{\frac{B(B - 1)}{2}}r\right|\right\},
\end{equation*}
and
\begin{equation*}
\bar{r}_\mathbb{M}(\kappa, h) = \inf\left\{r: \mathbb{E} \sup_{u \in \mathcal{F}_{h, r}} \left| \sqrt{\frac{2}{B(B - 1)}} \sum_{i < j} \sigma_{i, j}~u(X_i, X_j) \cdot h(X_i, X_j) \le \kappa \sqrt{\frac{B(B - 1)}{2}}r^2\right|\right\}.
\end{equation*}
While the difference on $r_E(\kappa, h)$ is only due to the double summation, the change in $\bar{r}_\mathbb{M}(\kappa, h)$ also comprises the removal of $Y$, as the general framework for pairwise learning does not involve any label. As a consequence, any $Y$ in the older definitions is replaced by $0$. Moreover, one can assess a $0$ noise $W$ such that $\|W\|_{L_2} = 0 \le 1 = \sigma$. This way, all $\sigma$'s encountered in older definitions and propositions can be replaced by $1$.
\end{definition}

\begin{lemma}\label{lem:small_ball}
For every $q > 2$ and $L \ge 1$, there are constants $B$ and $\kappa_0$ that depend only on $q$ and $L$ for which the following holds. If $\|h\|_{L_q} \le L \|h\|_{L_2}$ and $X_1, \ldots, X_B$ are independent copies of $X$, then
\begin{equation*}
\mathbb{P}\left\{\frac{2}{B(B - 1)} \sum_{1 \le i < j \le B} |h(X_i, X_j)| \ge \kappa_0 \|h\|_{L_2}\right\} \ge 0.9.
\end{equation*}
\end{lemma}
\proof{The proof is analogous to that of Lemma 3.4 in \citet{mendelson2017aggregation}, except that a version of Berry-Esseen theorem for $U$-statistics \citep{callaert1978berry} is used instead of the standard one.
}\qed

\begin{lemma}\label{lem:chebyshev}
For every $q > 2$ and $L \ge 1$, there is a constant $\kappa_1$ that depends only on $q$ and $L$ for which the following holds. If $X_1, \ldots, X_B$ are independent copies of $X$, then
\begin{equation*}
\mathbb{P}\left\{\frac{2}{B(B - 1)} \sum_{1 \le i < j \le B} |h(X_i, X_j)| \le \kappa_1 \|h\|_{L_2}\right\} \ge 0.9.
\end{equation*}
\end{lemma}
\proof{As $\left\{\frac{2}{B(B-1)} \sum_{i < j}|h(X_i, X_j)| \ge \kappa_1 \|h\|_{L_2}\right\} \subset \left\{\exists i < j,~|h(X_i, X_j)| \ge \kappa_1 \|h\|_{L_2}\right\}$, Chebyshev inequality gives
\begin{equation*}
\mathbb{P}\left\{\frac{2}{B(B - 1)} \sum_{1 \le i < j \le B} |h(X_i, X_j)| \ge \kappa_1 \|h\|_{L_2}\right\}
\le \frac{B(B-1)}{2}\mathbb{P}\left\{|h(X_i, X_j)| \ge \kappa_1 \|h\|_{L_2}\right\} \le \frac{B(B-1)}{2\kappa_1^2}.
\end{equation*}
Since $B$ only depends on $q$ and $L$ (see proof of \Cref{lem:small_ball}), so does $\kappa_1$.
}\qed

\begin{proposition}\label{prop:dist_oracle}
There are constants $\kappa, \eta, B, c >0$ and $0 < \alpha < 1 < \beta$ depending only on $q$ and $L$ for which the following holds. For a fixed $f^* \in \mathcal{F}$, let $r^* = \max\{\lambda_\mathbb{Q}(\kappa, \eta, H_{f^*}), r_E(\kappa, H_{f^*})\}$. For any $r \ge 2r^*$, with probability at least $1 - 2\exp(-cn)$, $\forall~H_f \in \mathcal{H}_\mathcal{F}$,
\begin{itemize}
\item If $\Phi_\mathcal{S}(f,{f^*}) \ge \beta r$, then $\beta^{-1}\Phi_\mathcal{S}(f, {f^*}) \le \|H_f - H_{f^*}\|_{L_2} \le \alpha^{-1}\Phi_\mathcal{S}(f, {f^*})$.
\item If $\Phi_\mathcal{S}(f, f^*) \le \beta r$, then $\|H_f - H_{f^*}\|_{L_2}\le (\beta/\alpha)r$.
\end{itemize}
\end{proposition}
\proof{Using \Cref{lem:small_ball} and \Cref{lem:chebyshev} with $h = H_f - H_{f^*}$, together with the union bound, it holds that for every block $\mathcal{B}_k$ one has with probability at least $0.8$
\begin{equation}\label{eq:encadrement}
\kappa_0 \|H_f - H_{f^*}\|_{L_2} \le \bar{U}_k(|H_f - H_{f^*}|)\le \kappa_1 \|H_f - H_{f^*}\|_{L_2}.
\end{equation}
Denoting by $I_k$ the indicator of this event, and by $\bar{I}_k$ its complementary, we have $\mathbb{E}[\bar{I}_k] \le 0.2$. Moreover,
\begin{equation*}
\mathbb{P}\left\{\sum_{k=1}^K I_k \ge 0.7K \right\} = 1 - \mathbb{P}\left\{\frac{1}{K}\sum_{k=1}^K \bar{I}_k \ge 0.3 \right\}.
\end{equation*}
When a MoU estimate is used, the $I_k$ are independent, since built on disjoint blocks, and the concentration of Binomial random variables allows to finish.
But interestingly, when a MoCU is used, it is straightforward to see that the last term is exactly the same quantity as the one involved in \eqref{eq:dev_split}. The same method can thus be used since an upper bound of $\mathbb{E}[\bar{I}_k]$ is already available. Precisely, choosing $\tau = 0.25 < 0.3$ and recalling $B = \lfloor n /K\rfloor$, it holds
\begin{equation*}
\mathbb{P}\left\{\frac{1}{K}\sum_{k=1}^K \bar{I}_k \ge 0.3 \right\} \le 2\exp\left(-2(0.05)^2K\right).
\end{equation*}
So the number of blocks which satisfy \eqref{eq:encadrement} is larger than $0.7K$ with probability at least $1 - 2\exp(-c_1K)$ for some positive constant $c_1$. The rest of the proof is similar to that of Proposition 3.2 in \citet{lugosi2016risk}.
}\qed

\begin{proposition}\label{prop:2}
Under the assumptions of \Cref{thm:tournament}, and using its notation, with probability at least
$$1 - 2\exp(-c_0n \min\{1, r^2\}),$$
$\forall~f \in \mathcal{F}$ if $\Phi_\mathcal{S}(f, f^*) \ge \beta r$ then $f^*$ defeats $f$. In particular $f^* \in H$, and $\forall~f \in H$, $\Phi_\mathcal{S}(f, f^*) = \|H_f - H_{f^*}\|_{L_2}\le \beta r$.
\end{proposition}

\proof{This proof carefully follows that of Proposition 3.5 in \citet{lugosi2016risk} (see Section 5.1 therein), so that only changes induced by pairwise objectives are detailed here. As discussed in \Cref{def:changes}, every $Y$ can be replaced by $0$, and every $\sigma$ by $1$. Attention must also be paid to the fact that in the context of means, $m$, the cardinal of each block, is also equal to $\binom{m}{1}$, the number of possible $1$-combinations. In our notation, it thus may sometimes be identified to $B$, the cardinal of the blocks, and sometimes to $\frac{B(B - 1)}{2}$, the number of pairs per block.

{\it Proof of pairwise Lemma 5.1} First, one may rewrite
\begin{equation*}
\mathbb{Q}_{f, g} = \frac{2}{B(B - 1)} \sum_{i < j}\left(H_f(X_i, X_j) - H_g(X_i, X_j)\right)^2,
\end{equation*}
\begin{equation*}
\mathbb{M}_{f, g} = \frac{4}{B(B - 1)} \sum_{i < j}\left(H_f(X_i, X_j) - H_g(X_i, X_j)\right) \cdot H_g(X_i, X_j),
\end{equation*}
and
\begin{equation*}
R_k(u, t) = \left|\{(i, j) \in \mathcal{B}_k^2 : i < j, |u(X_i, X_j)| \ge t\}\right| = \sum_{i < j \in \mathcal{B}_k^2} \mathbbm{1}\{|u(X_i, X_j) \ge t|\}.
\end{equation*}
Since all pairs are not independent, even if the $X_i$'s are, one cannot use directly the proposed method. Instead, the Hoeffding inequality for $U$-statistics gives that the probability of each $R_k(H_f - H_{f^*}, \kappa_0 r)$ to be greater than $\frac{B(B - 1)\rho_0}{4}$ is greater than $1 - \exp(- \frac{B\rho_0^2}{4})$. For $\tau$ small enough, we still have that this probability is greater than $1 - \tau / 12$. Aggregating the Bernoulli may be done in two ways. If we deal with a MoU estimate, the independence between blocks leads to the same conclusion. If a MoRU estimate is used instead, the remark made
for \Cref{prop:dist_oracle}'s proof is again valid, and one can conclude.

The next difficulty arises with the bounded differences inequality for $\Psi$. If a MoU estimate is used, changing one sample $X_i'$ only affects one block, and generates a $1/K$ difference at most, exactly like with MoM, so that the bound holds the same way. On the contrary, if a MoRU is used, there is no guarantee that the replaced sample contaminates all $K$ blocks. The analysis of the MoRU behavior in that case is a bit trickier, and we restrict ourselves to MoU estimates for the matches.

The end of the proof uses a standard symmetrization argument.
This kind of arguments still apply to $U$-statistics (see \textit{e.g.} p.150 of  \citet{pena1999decoupling}), and the proof is completed in the pairwise setting.

\begin{samepage}
{\it Proof of pairwise Lemma 5.2}
\begin{align*}
\mathbb{P}\left\{\left|\frac{2}{B(B - 1)} \sum_{i < j} U_{i, j} - \mathbb{E}U\right| \ge t \right\} &\le \frac{2}{B(B - 1)t} \mathbb{E}\left|\sum_{i < j} U_{i, j} - \mathbb{E}U\right|\\
&\le \frac{2}{B(B - 1)t} \sqrt{\mathbb{E}\left|\sum_{i < j} U_{i, j} - \mathbb{E}U\right|^2}\\
&\le \frac{2}{B(B - 1)t} \sqrt{\sum_{i < j,~k < l} \mathbb{E}[U_{i, j}U_{j, k}] - (\mathbb{E}U)^2}\\
&\le \frac{2}{B(B - 1)t} \sqrt{ \frac{B(B - 1)}{2}\left(\mathbb{E}[U^2] - \left(\mathbb{E}U\right)^2\right) + B(B - 1)(B - 2)\sigma_1^2)}\\
&\le \frac{\sqrt{2 (B-2)}}{\sqrt{B(B - 1)}t}\|U\|_{L_2}\\
\mathbb{P}\left\{\left|\frac{2}{B(B - 1)} \sum_{i < j} U_{i, j} - \mathbb{E}U\right| \ge t \right\} &\le \frac{\sqrt{2}}{\sqrt{B}t}\|U\|_{L_2}
\end{align*}
\end{samepage}

After that, every case needing a pairwise investigation has already been treated earlier in the section: Binomial concentration, bounded differences inequality, symmetrization arguments. So is \Cref{prop:2} proved.

{\bf \Cref{thm:tournament}'s proof.} \Cref{prop:dist_oracle} and \Cref{prop:2} gives that if any $f \in \mathcal{F}$ wins all its matches, then with probability at least $1 - 2\exp(-c_0n \min\{1, r^2\})$ $\|H_f - H_{f^*}\|_{L_2} \le cr$. Hence it also holds with the same probability:
$$
\mathcal{R}(f) - \mathcal{R}(f^*) = \|H_f\|_{L_2} - \|H_{f^*}\|_{L_2} \le \|H_f - H_{f^*}\|_{L_2} \le cr.
$$
\qed

Although this extension to the pairwise learning framework deals with any loss $\ell$, it is important to notice that the extension of Theorem 2.11 in \citet{lugosi2016risk} is applied to $H_f$, and not $f$ directly.
$H_f$ is penalized via the quadratic loss (as in \citet{lugosi2016risk}), so that the Theorem apply, up to technicalities induced by the $U$-statistics.
Doing so, one achieved a control on $\|H_f - H_f^*\|_{L_2}$ (as in \citet{lugosi2016risk}), which is equal to $\|H_f - H_{f^*}\|_{L_2}$ thanks to the remark stated in the first paragraph of Subsection 3.3.
This quantity happens to be greater than the excess risk of $f$, hence the conclusion.
Formally, the tournament procedure outputs a $\hat{H}_f$, and one has to recover the $\hat{f}$ such that $\hat{H}_f = H_{\hat{f}}$.
Knowing the dependence between $f$ and $H_f$, and with the ability to evaluate $\hat{H}_f$, which is known, on any pair, this last step should not be too difficult.\\

About the extension to pairwise learning, one should keep in mind that the general framework does not involve any target $Y$.
Instead, one seeks directly to minimize $\ell(f, X, X') = \sqrt{\ell(f, X, X')}^2 = (\sqrt{\ell(f, X, X')} - 0)^2 = (H_f(X, X') - 0)^2$.
We recover the setting of Theorem 2.11 in \citet{lugosi2016risk}: quadratic loss, with $Y=0$, for the decision function $H_f$.
The only novelty to address here is the fact that $H_f$ depends on two random variables $X$ and $X'$.
And this is precisely what has been done in this subsection.

\clearpage

\section{More Numerical Results}

\subsection{MoRM Estimation Results}

\begin{table*}[h!]
\caption{Quadratic Risks for the Mean Estimation, $\delta = 0.001$}
\label{tab:MoRM_QR_full}
\vskip 0.15in
\begin{center}
\begin{small}
\begin{sc}
\begin{tabular}{lcccc}
\toprule
                                  & Normal $(0, 1)$       & Student $(3)$         & Log-normal $(0, 1)$   & Pareto $(3)$\\\midrule

MoM                               & 0.00149 $\pm$ 0.00218 & 0.00410 $\pm$ 0.00584 & 0.00697 $\pm$ 0.00948 & \textbf{1.02036 $\pm$ 0.06115}\\
$\text{MoRM}_\text{$1/6$, SWoR}$  & 0.01366 $\pm$ 0.01888 & 0.02947 $\pm$ 0.04452 & 0.06210 $\pm$ 0.07876 & 1.12256 $\pm$ 0.14970\\
$\text{MoRM}_\text{$1/6$, MC}$    & 0.01370 $\pm$ 0.01906 & 0.02917 $\pm$ 0.04355 & 0.06167 $\pm$ 0.07143 & 1.13058 $\pm$ 0.14880\\
$\text{MoRM}_\text{$3/10$, SWoR}$ & 0.00255 $\pm$ 0.00361 & 0.00602 $\pm$ 0.00868 & 0.01241 $\pm$ 0.01610 & 1.05458 $\pm$ 0.07041\\
$\text{MoRM}_\text{$3/10$, MC}$   & 0.00264 $\pm$ 0.00372 & 0.00622 $\pm$ 0.00895 & 0.01283 $\pm$ 0.01650 & 1.05625 $\pm$ 0.07298\\
$\text{MoRM}_\text{$9/20$, SWoR}$ & 0.00105 $\pm$ 0.00148 & \textbf{0.00264 $\pm$ 0.00372} & \textbf{0.00497 $\pm$ 0.00668} & 1.02802 $\pm$ 0.04903\\
$\text{MoRM}_\text{$9/20$, MC}$   & \textbf{0.00105 $\pm$ 0.00146} & 0.00265 $\pm$ 0.00374 & 0.00499 $\pm$ 0.00673 & 1.02985 $\pm$ 0.04880\\\bottomrule
\end{tabular}
\end{sc}
\end{small}
\end{center}
\vskip -0.1in
\end{table*}

\begin{figure*}[!h]
\centering
\subfigure{\includegraphics[width=0.45\textwidth]{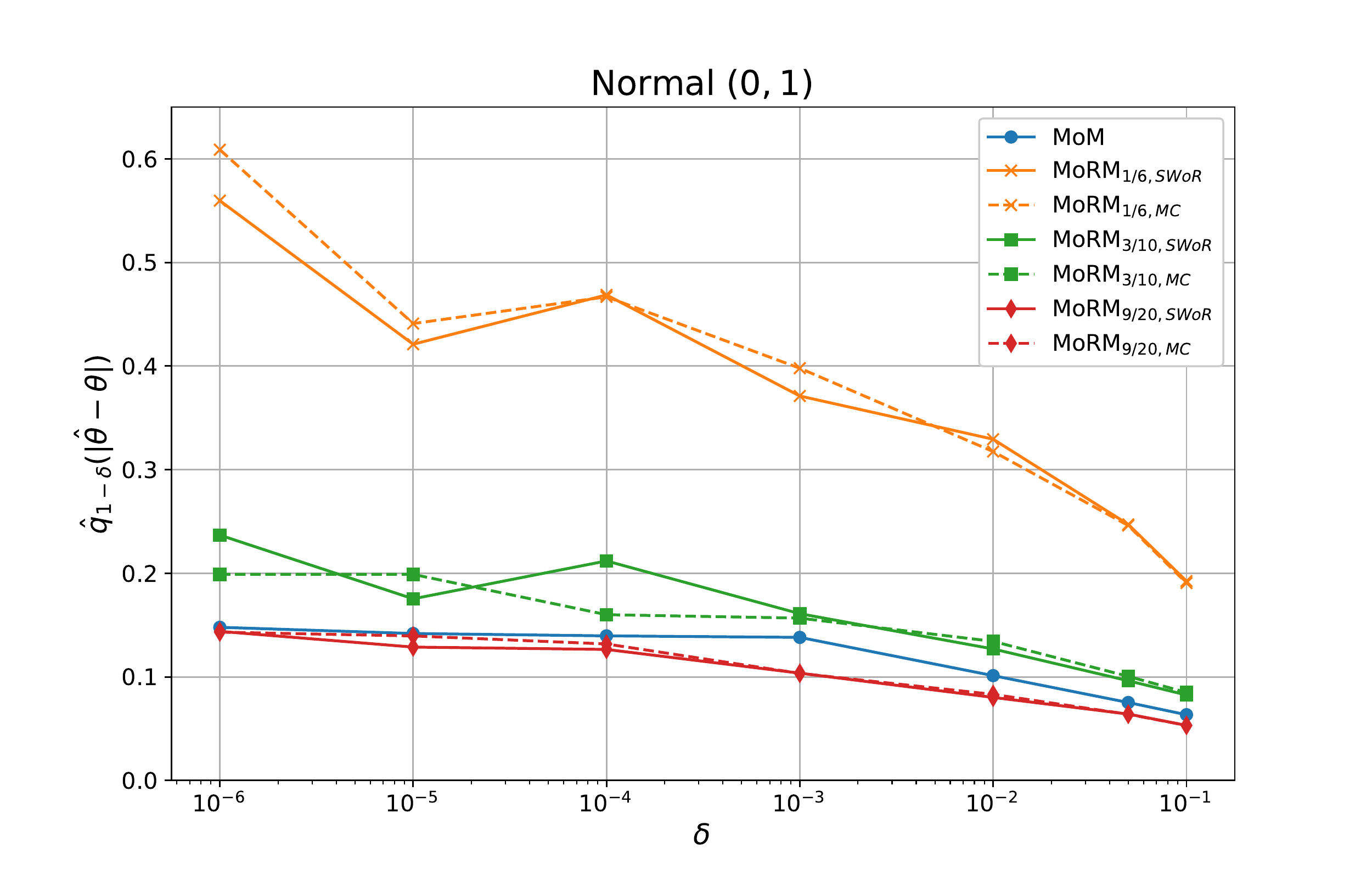}}
\subfigure{\includegraphics[width=0.45\textwidth]{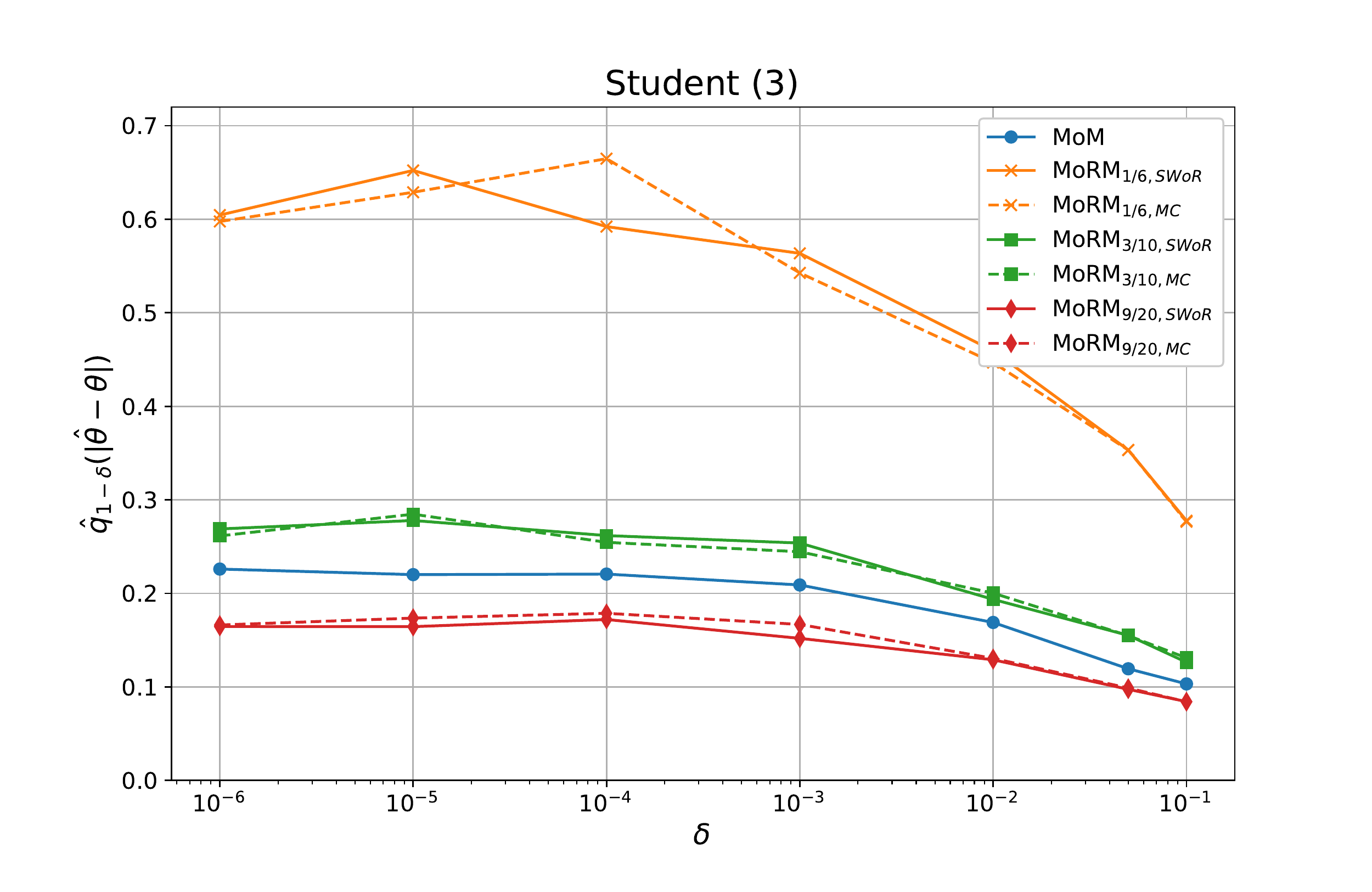}}\\
\subfigure{\includegraphics[width=0.45\textwidth]{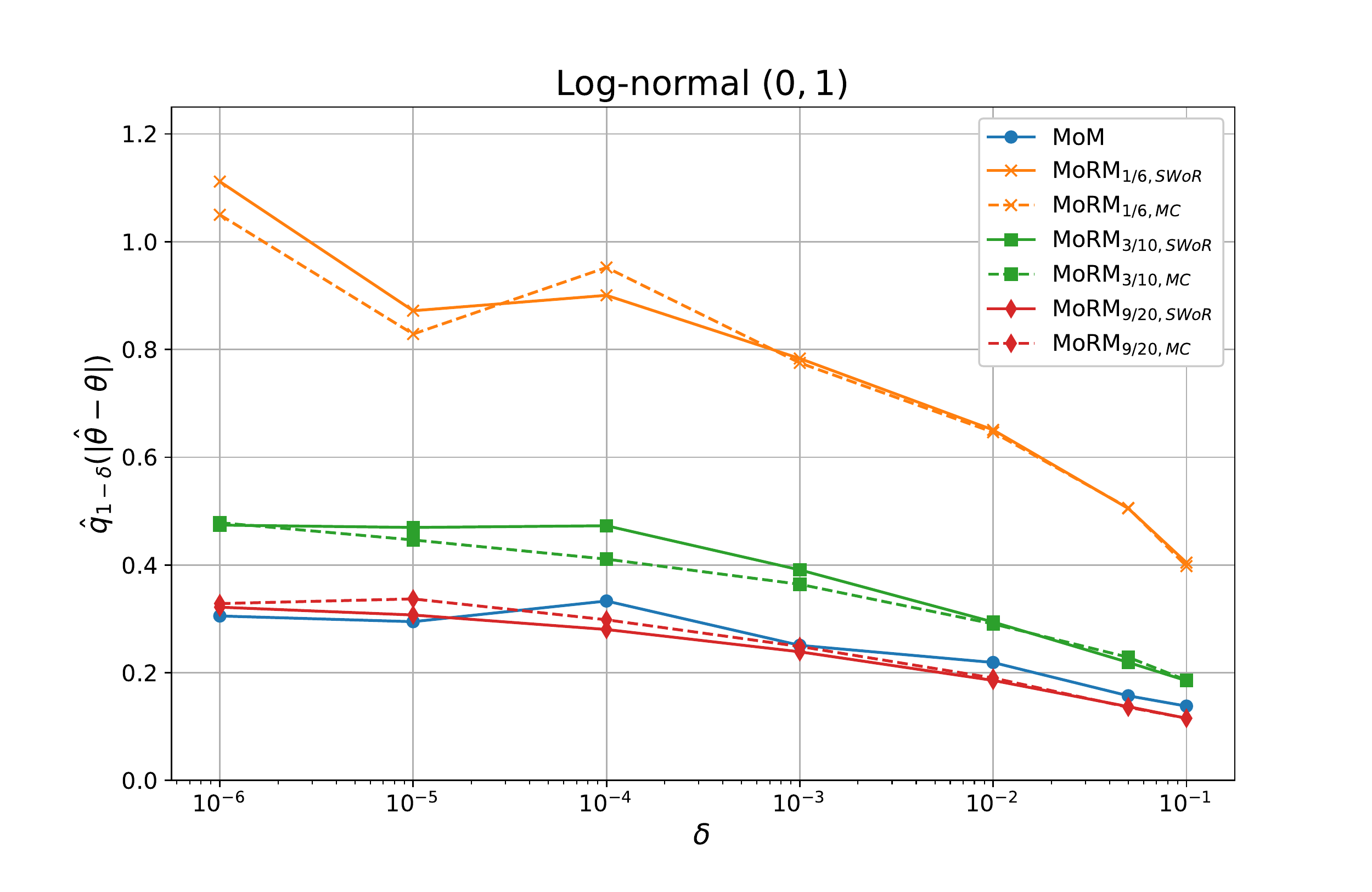}}
\subfigure{\includegraphics[width=0.45\textwidth]{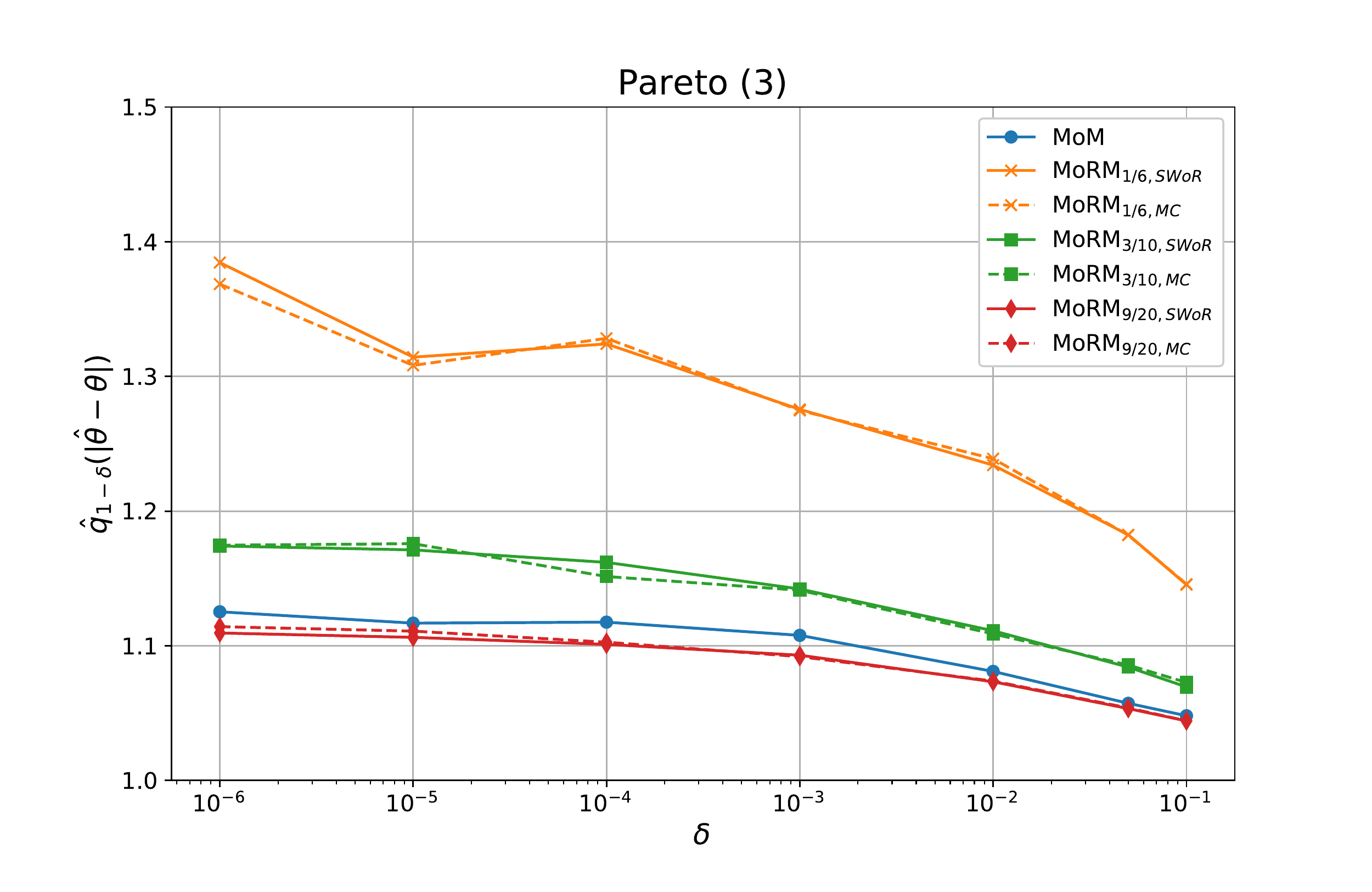}}
\caption{Empirical Quantiles for the Different Mean Estimators on 4 Laws}
\label{fig:quantiles}
\end{figure*}

The empirical quantiles confirm the quadratic risks results: the $\tau$ parameter is crucial, making MoRM the worst or the best estimate depending on its value. The sampling scheme does not affect to much the performance, even if the MC scenario is much more complex to analyze theoretically.

\newpage

\subsection{MoRU Estimation Results}

\begin{table*}[h!]
\caption{Quadratic Risks for the Variance Estimation, $\delta = 0.001$}
\label{tab:MoIU_QR_full}
\vskip 0.15in
\begin{center}
\begin{small}
\begin{sc}
\begin{tabular}{lcccc}
\toprule
                                  & Normal $(0, 1)$       & Student $(3)$         & Log-normal $(0, 1)$   & Pareto $(3)$\\\midrule
$\text{MoU}_\text{1/2; 1/2}$      & 0.00409 $\pm$ 0.00579 & 1.72618 $\pm$ 28.3563 & 2.61283 $\pm$ 23.5001 & 1.35748 $\pm$ 36.7998\\
$\text{MoU}_\text{Partition}$     & 0.00324 $\pm$ 0.00448 & \textbf{0.38242 $\pm$ 0.31934} & \textbf{1.62258 $\pm$ 1.41839} & \textbf{0.09300 $\pm$ 0.05650}\\
$\text{MoRU}_\text{SWoR}$         & 0.00504 $\pm$ 0.00705 & 0.51202 $\pm$ 3.88291 & 2.01399 $\pm$ 4.85311 & 0.09703 $\pm$ 0.07116\\
$\text{MoIU}_\text{$1/6$, SWoR}$  & 0.00206 $\pm$ 0.00285 & 1.78161 $\pm$ 34.7216 & 2.50529 $\pm$ 21.8989 & 1.37800 $\pm$ 40.1308\\
$\text{MoIU}_\text{$1/6$, MC}$    & \textbf{0.00205 $\pm$ 0.00281} & 1.65481 $\pm$ 26.2157 & 2.61701 $\pm$ 24.7918 & 1.50578 $\pm$ 42.9135\\
$\text{MoIU}_\text{$3/10$, SWoR}$ & 0.00216 $\pm$ 0.00301 & 1.13887 $\pm$ 16.9511 & 2.07136 $\pm$ 14.8312 & 0.85041 $\pm$ 21.9916\\
$\text{MoIU}_\text{$3/10$, MC}$   & 0.00211 $\pm$ 0.00288 & 1.22402 $\pm$ 17.4715 & 2.16590 $\pm$ 15.2378 & 0.89035 $\pm$ 22.2866\\\bottomrule
\end{tabular}
\end{sc}
\end{small}
\end{center}
\vskip -0.1in
\end{table*}

\begin{figure*}[!h]
\centering
\subfigure{\includegraphics[width=0.45\textwidth]{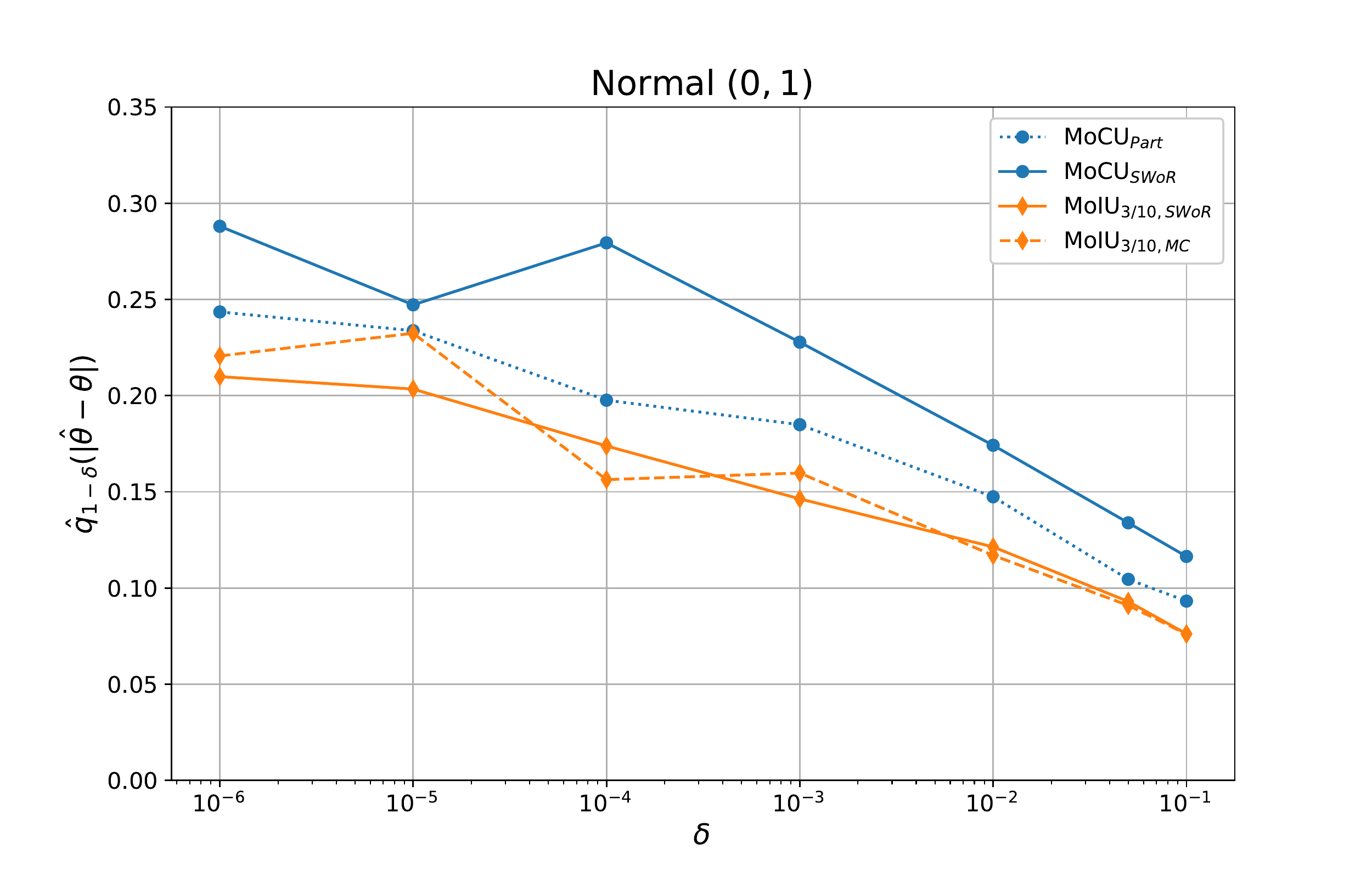}}
\subfigure{\includegraphics[width=0.45\textwidth]{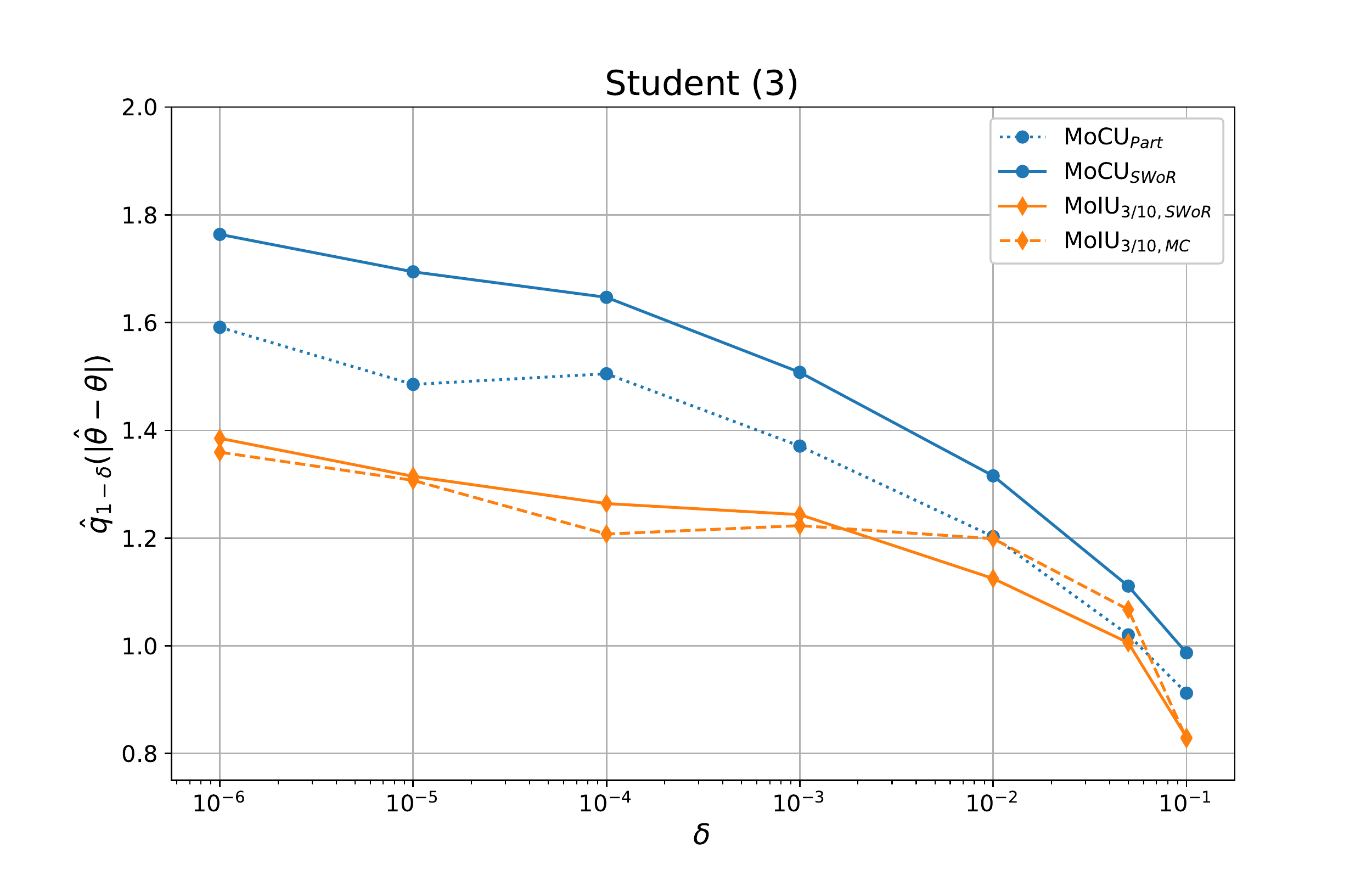}}\\
\subfigure{\includegraphics[width=0.45\textwidth]{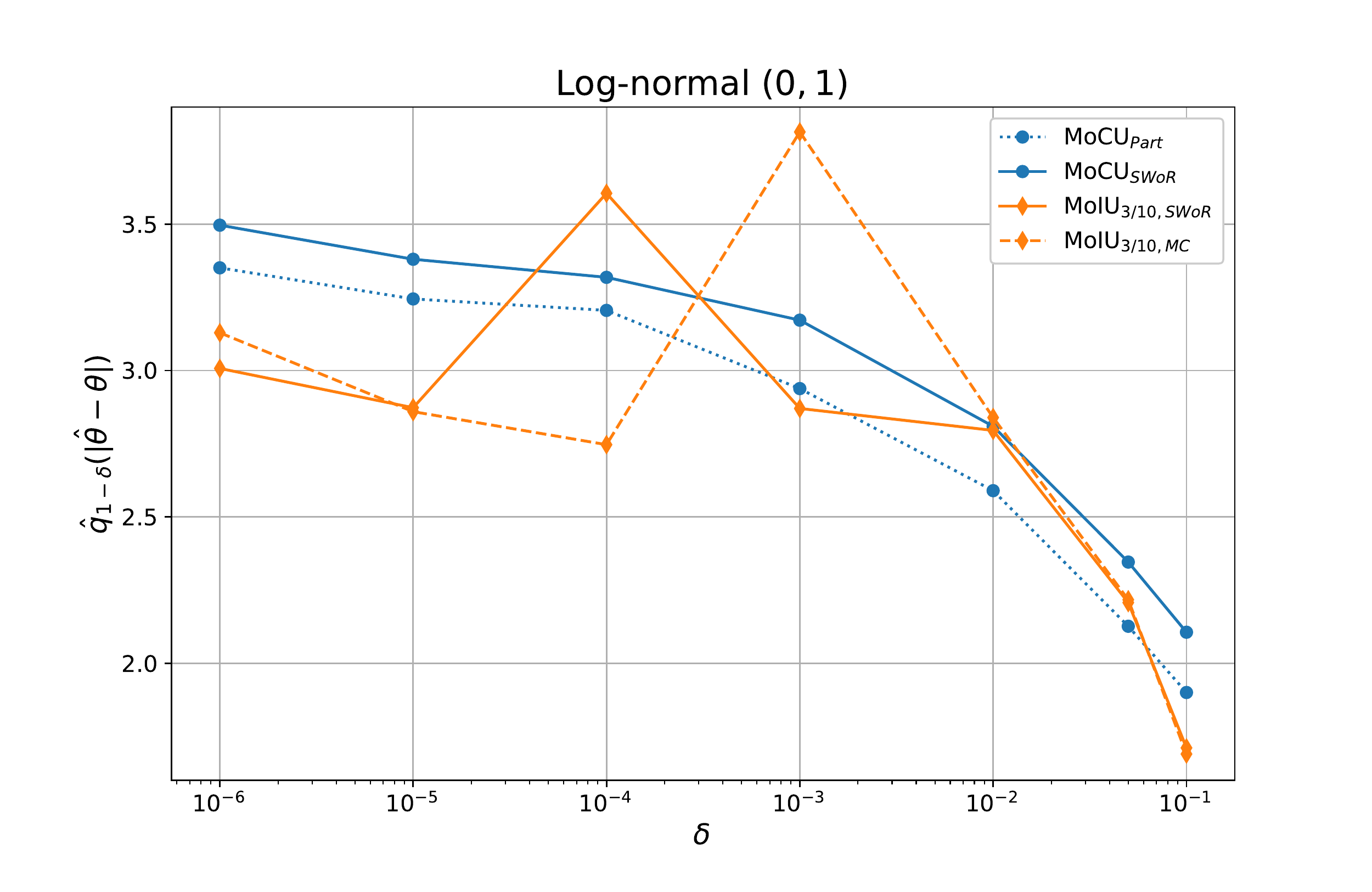}}
\subfigure{\includegraphics[width=0.45\textwidth]{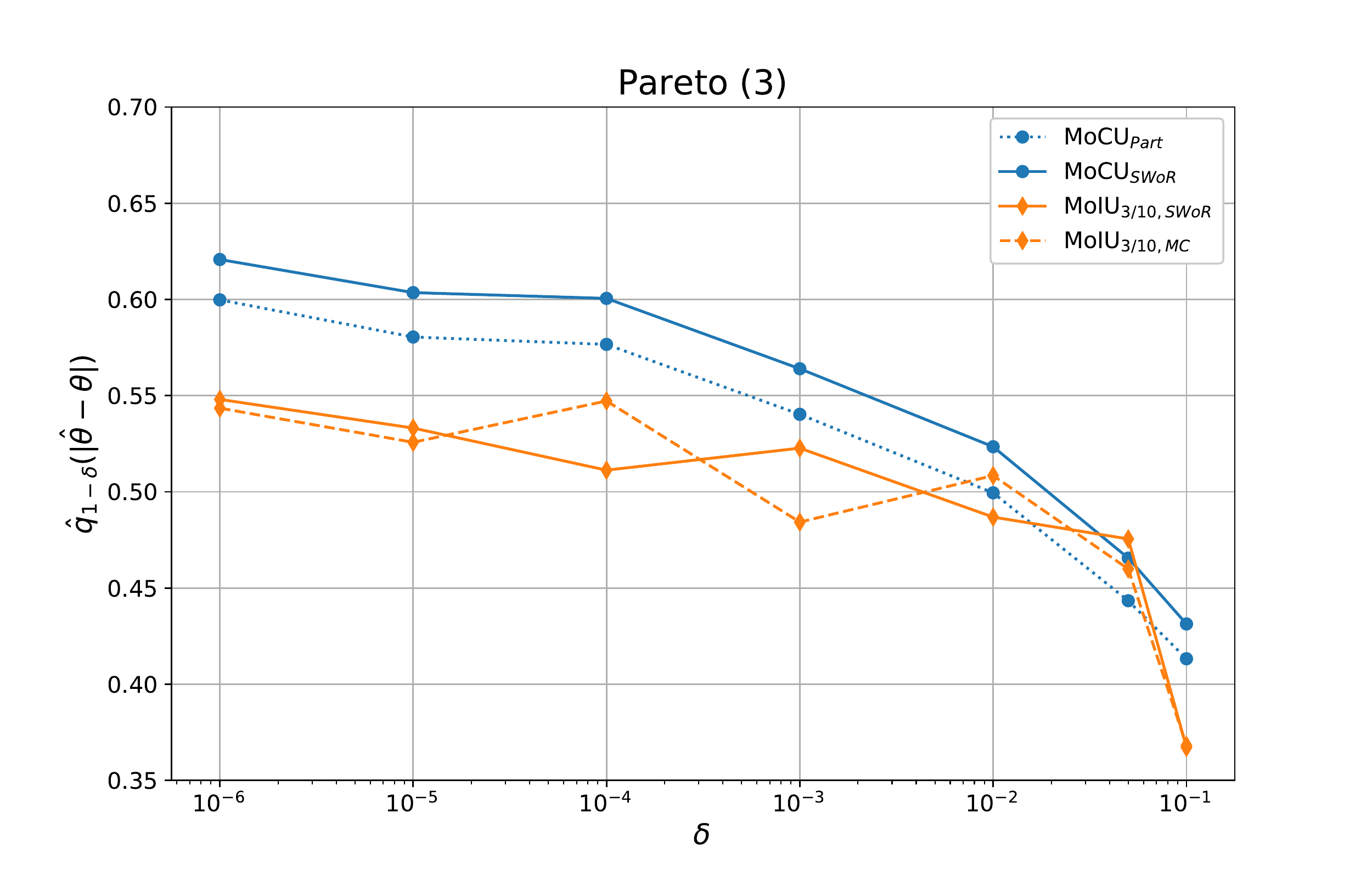}}
\caption{Empirical Quantiles for the Different Variance Estimators on 4 Laws}
\label{fig:Uquantiles}
\end{figure*}

The partitioning MoU seems to outperform every other estimate. One explanation can be that an extreme value may \textit{corrupt} only one block within this method, whereas randomized versions can suffer from it in several blocks.

\newpage

\subsection{MoRU Learning Results: a Metric Learning Application}

\begin{figure*}[!h]
\centering
\subfigure{\includegraphics[width=0.73\textwidth]{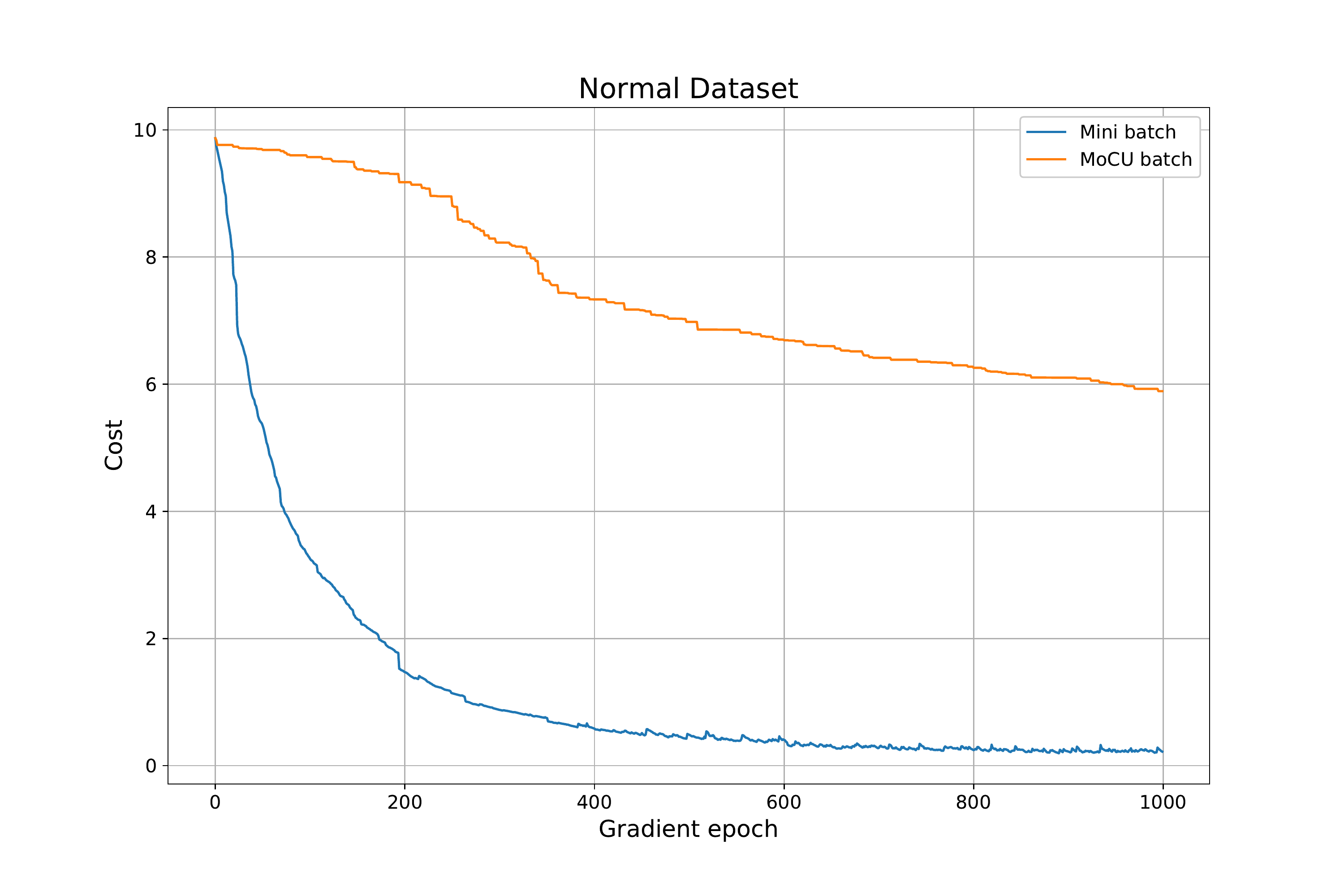}}\\
\subfigure{\includegraphics[width=0.73\textwidth]{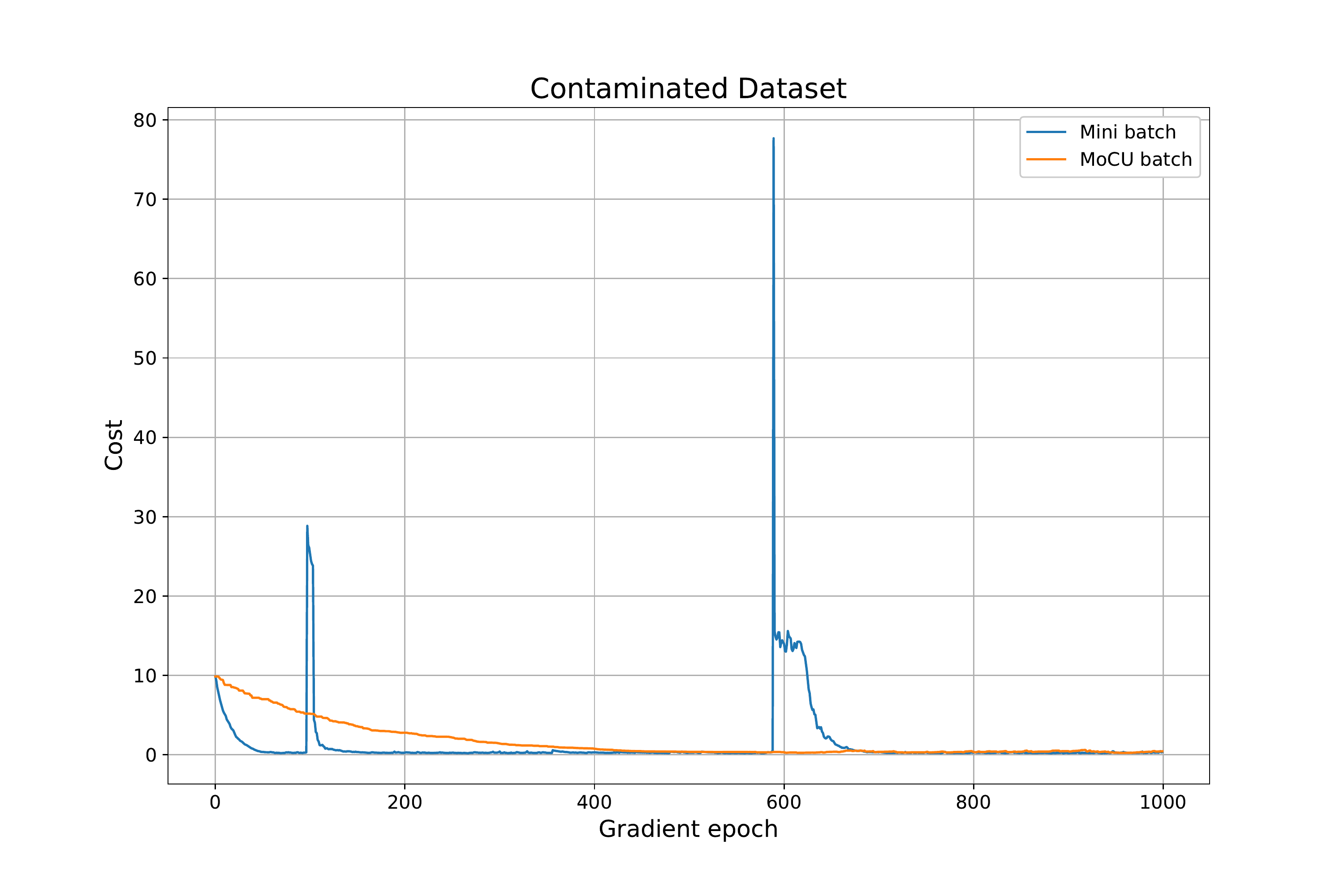}}
\caption{Gradient Descent Convergences for Normal and Contaminated Datasets}
\label{fig:Uquantiles2}
\end{figure*}

In this experiment, we try to learn from points that are known to be close or not, a distance that fits them, over the set of all possible Mahalanobis distances, \textit{i.e.} $d(x,y) = \sqrt{(x-y)^\top M (x -y)}$ for some positive definite matrix $M$.
It is done through mini-batch gradient descent over the parameter $M$.
On the normal iris dataset (top), we see that standard mini-batches perform well, while MoRU mini-batches induce a much slower convergence.
But in the spirit of \citet{lecue2017robust}, the experiment is also run on a (artificially) corrupted dataset (bottom).
This highlights how well can the MoM-like estimators behave in the presence of outliers.
Indeed, although the convergence with the MoRU mini-batches remains slower than with the standard ones on the normal regime, they avoid peaks, presumably caused by the presence of one (or more) outlier in the mini-batch.
This makes it a very interesting alternative in the context of highly corrupted data.
Experiments run on the same dataset for clustering purposes show the same behavior.

\newpage

\section{Alternative Sampling Schemes - Possible Extensions}

As pointed out in Remark \ref{rk:alt1} and in the discussion in the end of Section \ref{sec:main}, the approaches investigated in the present paper could be implemented with sampling procedures different from the SRSWoR scheme. It is the purpose of this section to review possible alternatives and discuss the technical difficulties inherent to the study of their performance.

\subsection{MoRM using a Sampling with Replacement Scheme}\label{app:morm}

Rather than forming $K\geq 1$ data blocks of size $B\leq n$ from the original sample $\mathcal{S}_n$ by means of a SRSWoR scheme, one could use a Monte-Carlo procedure and draw independently $K$ times an arbitrary number $B$ of observations with replacement. In this case, each data block $\mathcal{B}_k$ is characterized by a random vector ${\bm e}_k=(e_{k,1},\; \ldots,\; e_{k,B})$ independent from $\mathcal{S}_n$, where, for each draw $b\in\{1,\; \ldots,\; B \}$, $e_{k,b}=(e_{k,b}(1),\; \ldots,\; e_{k,b}(n))$ is a multinomial random vector in $\{0,1\}^n$ indicating the index of the observation randomly selected: for any $i\in\{1,\; \ldots,\; n \}$, $e_{k,b}(i)=1$ if $i$ has been chosen, $e_{k,b}(i)=0$ otherwise and $\mathbb{P}\{e_{k,b}(i)=1 \}=1/n$ (notice also that $\sum_{i=1}^ne_{k,b}(i)=1$ with probability one). In this case, the empirical mean based on block $\mathcal{B}_k$ can be written as
$$
\tilde{\theta}_k=\frac{1}{B}\sum_{b=1}^B\langle e_{k,b},\; {\bf Z}_n  \rangle,
$$
where  ${\bf Z}_n=(Z_1,\; \ldots,\; Z_n)$ and $\langle .,\; . \rangle$ is the usual Euclidean scalar product on $\mathbb{R}^n$. Conditioned upon $\mathcal{S}_n$, the $\tilde{\theta}_k$'s are i.i.d. and we have $\mathbb{E}[ \tilde{\theta}_1 \mid \mathcal{S}_n ]= \hat{\theta}_n$, as well as
$$
\text{Var}(\tilde{\theta}_1 | \mathcal{S}_n) =\frac{1}{B}\left(\frac{1}{n}\sum_{j=1}^n Z_j^2 - \hat{\theta}_n^2\right).
$$
 The corresponding variant of the MoM estimator is
$$
\mormbis=\text{median}\left(\tilde{\theta}_1,\; \ldots,\; \tilde{\theta}_K  \right).
$$

Observe that this estimation procedure offers a greater flexibility, insofar as both $K$ and $B$ can be arbitrarily chosen. In addition, the variance of the block estimators $\tilde{\theta}_k$:
$$
\text{Var}(\tilde{\theta}_1)=\text{Var}\left( \mathbb{E}[ \tilde{\theta}_1 \mid \mathcal{S}_n ] \right)+\mathbb{E}\left[\text{Var}\left( \tilde{\theta}_1 \mid \mathcal{S}_n \right)  \right]=\left(\frac{1}{n}+\frac{1}{B}\left(1-\frac{1}{n}  \right)\right)\sigma^2.
$$
It is comparable to that of the $\bar{\theta}_k$'s for the same block size $B\leq n$, although always larger: $\text{Var}(\tilde{\theta}_1)-\text{Var}(\bar{\theta}_1)=(1-1/B)\sigma^2/n\geq 0$.
However, investigating the accuracy of $\mormbis$ is challenging, due to the fact that it is far from straightforward to study the concentration properties of the random quantity
$$
\tilde{U}_n^{\varepsilon}=\mathbb{P}\{\vert \tilde{\theta}_1- \theta \vert>\varepsilon \mid \mathcal{S}_n \}.
$$
Even if Chebyshev's inequality yields
$$
\mathbb{E}[\tilde{U}_n^{\varepsilon} ]=p^{\varepsilon}\leq \left(\frac{1}{n}+\frac{1}{B}\left(1-\frac{1}{n}  \right)\right)\frac{\sigma^2}{\epsilon^2},$$
the r.v. $\tilde{U}_n^{\varepsilon}$ is a complex functional of the original data $\mathcal{S}_n$, due to the possible multiple occurrence of a given observation in a single sample obtained through sampling with replacement. In particular, in contrast to $U_n^{\varepsilon}$, it is not a $U$-statistic of degree $B$. Viewing it as a function of the $n$ i.i.d. random variables $Z_1,\; \ldots,\; Z_n$ and observing that changing the value of any of them can change its value by at most $1-(1-1/n)^{B}\leq B/n$, the bounded difference inequality (see \citet{mcdiarmid_1989}) gives only: $\forall t>0$,
\begin{equation}\label{eq:bound_Mc}
\mathbb{P}\left\{  \tilde{U}_n^{\varepsilon} - p^{\varepsilon} \geq t  \right\}\leq \exp\left( -2\frac{n}{B^2}t^2 \right),
\end{equation}
while we have $\mathbb{P}\{ U_n^{\varepsilon} - p^{\varepsilon} \geq t \}\leq \exp( -2(n/B)t^2 )$. Hence, the bound \eqref{eq:bound_Mc} is not sharp enough to establish guarantees for the estimator $\mormbis$ similar to those stated in Proposition \ref{prop:morm}.

\subsection{Medians-of-Incomplete $U$-statistics}

 The computation of the $U$-statistic \eqref{eq:Ustat} is expensive in the sense that it involves the summation of $\mathcal{O}(n^2)$ terms.
The concept of \textit{incomplete $U$-statistic}, see \citet{Blom76} permits to address the computational issue raised by the expensive calculation of the $U$-statistic \eqref{eq:Ustat}, which involves the summation of $\mathcal{O}(n^2)$ terms, so as to achieve a trade-off between scalability and variance reduction.
In one of its simplest forms, it consists in selecting a subsample of size $M \geq $ by \textit{sampling with replacement} (\textit{i.e.} Monte-Carlo scheme) in the set $
\Lambda=\{(i,j):\; 1\leq i<j\leq n \},
$ of all pairs of observations that can be formed from the original sample.
Denoting by $\{(i_1,j_1),\; \ldots,\; (i_M,j_M)\}\subset \Lambda$ the subsample thus drawn, the incomplete version of the $U$-statistic \eqref{eq:Ustat} is:
\begin{equation*}\label{eq:incomp_Ustat}
\widetilde{U}_M(h)=\frac{1}{M}\sum_{m=1}^M h(X_{i_m},X_{j_m}).
\end{equation*}
As can be easily shown, \eqref{eq:incomp_Ustat} is an unbiased estimator of $\theta$. More precisely, its conditional expectation given $\mathcal{S}_n$ is equal to $U_n(h)$ and its variance is
$$
\text{Var}\left( \widetilde{U}_M(h) \right)=\left(1-\frac{1}{M}\right)\text{Var}\left( U_n(h)  \right)+\frac{\sigma^2(h)}{M}.
$$

Repeating independently the sampling procedure $K$ times in order to compute incomplete $U$-statistics $\widetilde{U}_{M,1}(h),\; \ldots,\; \widetilde{U}_{M,K}(h)$ (conditionally independent given the original dataset $\mathcal{S}_n$), one may consider the Median of Incomplete $U$-statistic:
\begin{equation*}
\moiu=\text{median}\left(\widetilde{U}_{M,1}(h),\; \ldots,\; \widetilde{U}_{M,K}(h) \right).
\end{equation*}
Although the variance of the $\widetilde{U}_{M,k}(h)$'s are smaller than that of the $\hat{U}_k(h)$'s (and that of the $\bar{U}_k(h)$'s) for the same number of pairs involved in the computation of each estimator, the argument underlying Proposition \ref{prop:U_stat_rand}'s proof cannot be adapted to the present situation because the concentration properties of the random quantity
\begin{equation*}\label{eq:Wbis}
\widetilde{W}_n^{\varepsilon}=\mathbb{P}\left\{ \left\vert \widetilde{U}_M(h) -\theta(h) \right\vert>\varepsilon  \mid \mathcal{S}_n \right\},
\end{equation*}
which has mean
$$
\widetilde{q}^{\varepsilon}=\mathbb{P}\left\{ \left\vert \widetilde{U}_M(h) -\theta(h) \right\vert>\varepsilon  \right\}\leq \frac{\text{Var}\left( \widetilde{U}_M(h) \right)}{\varepsilon^2}=\mathcal{O}\left(\frac{1}{M\varepsilon^2}\right),
$$
are very difficult to study, due to the complexity of the data functional \eqref{eq:Wbis}. Like in SM \ref{app:morm}, a straightforward application of the bounded difference inequality gives: $\forall t>0$,
$$
\mathbb{P}\left\{ \widetilde{W}_n^{\varepsilon} - \widetilde{q}^{\varepsilon}\geq t \right\}\leq \exp\left(  -2\frac{n}{M^2}t^2\right).
$$
Obviously, this bound is not sharp enough to yield a bound of the same order as those in Proposition \ref{prop:U_stat_comp} and Proposition \ref{prop:U_stat_rand}.
\medskip

\noindent{\bf Alternative sampling schemes.} Other procedures than the Monte-Carlo scheme above  can of course be considered to build a subset of all pairs of observations and compute an incomplete $U$-statistic. Refer to \citet{Lee90} or to subsection 3.4 in \citet{CCB16} for further details. When selecting a subsample of size $M \leq  n(n-1)/2$ by \textit{simple random sampling without replacement} (SRSWoR) in the set of all pairs of observations that can be formed from the original sample, the version of \eqref{eq:incomp_Ustat}obtained has conditional expectation and variance:
\begin{eqnarray*}
\mathbb{E}\left[ \widetilde{U}_M(h)  \mid X_1,\; \ldots,\; X_n\right]&=&U_n(h),\\
\var{\widetilde{U}_M(h) \mid X_1,\; \ldots,\; X_n}&=&\frac{\binom{n}{2}-M}{\binom{n}{2}}\times \frac{\hat{V}^2_n(h)}{M},
\end{eqnarray*}
where
$$
\hat{V}^2_n(h)=\frac{1}{\binom{n}{2}-1}\sum_{i<j}\left( h(X_i,X_j)-U_n(h)\right)^2.
$$
Hence, its variance can be expressed as
\begin{equation*}
\var{\widetilde{U}_M(h)} =~ \var{U_n(h)} ~+
\frac{\binom{n}{2} - M}{\binom{n}{2}-1}\times \frac{1}{M}\left( \sigma^2(h)+\var{U_n(h)} \right)=\mathcal{O}(1/M).
\end{equation*}

In contrast with the situation investigated in subsection \ref{subsec:morm}, the conditional expectation $\widetilde{W}_{n}^{\varepsilon} $ cannot be viewed as a $U$-statistic of degree $M$, insofar as the $n(n-1)/2$ variables $\{h(X_i,X_j):\; i<j \}$ are not i.i.d. r.v.'s.
Here as well, the bounded difference inequality is not sufficient to get bounds of the same order as that obtained in those in Proposition \ref{prop:U_stat_comp} and Proposition \ref{prop:U_stat_rand}, jumps
\begin{equation*}\label{eq:jump_bound}
1-\binom{(n-1)(n-2)/2}{M}/\binom{n(n-1)/2}{M}
\end{equation*}
being at least of order $M/n$, as Stirling's formula shows.